%% file: iclr2026_conference.tex
\documentclass{article} 

\usepackage{iclr2026_conference,times}
\usepackage[hyphens]{url}
\usepackage{hyperref}
\hypersetup{breaklinks=true}

\input{math_commands.tex}

\iclrfinalcopy 
\usepackage{hyperref}
\usepackage{url}
\usepackage{booktabs} 
\usepackage{colortbl} 
\usepackage{tcolorbox}
\usepackage{makecell}
\usepackage{adjustbox}
\usepackage{soul}
\usepackage{graphicx}
\usepackage{cleveref}
\usepackage{array}
\usepackage{adjustbox}
\usepackage{tcolorbox}
\usepackage{booktabs}
\usepackage[caption=false]{subfig}
\usepackage{makecell}
\usepackage{multirow}
\usepackage{tabularx}
\usepackage{caption}
\usepackage{amsmath}
\usepackage{subcaption}
\usepackage{enumitem}
\usepackage{wrapfig}
\usepackage{longtable}
\usepackage{tabto}
\usepackage{hyperref}
\usepackage{listings}
\usepackage{subcaption}
\usepackage{tcolorbox}
\usepackage{wrapfig}
\usepackage{graphicx}
\usepackage{makecell}
\usepackage{fontawesome}
\usepackage{hyperref}
\usepackage{url}

\AtBeginDocument{%
  }

\title{Menta: A Small Language Model for On-Device Mental Health Prediction}

\author{
\parbox{\linewidth}{
Tianyi Zhang$^{1}$\thanks{These authors contributed equally to this work.},
Xiangyuan Xue$^{2}\footnotemark[1]$,
Lingyan Ruan$^1$, Shiya Fu$^1$, Feng Xia$^3$, Simon D'Alfonso$^1$,\\
Vassilis Kostakos$^1$, Ting Dang$^1$, Hong Jia$^2$
}
\AND
\parbox{\linewidth}{
$^1$The University of Melbourne \quad
$^2$The University of Auckland \quad
$^3$RMIT University \\
}
}

\begin{document}

\definecolor{locations}{RGB}{214, 233, 213} 
\definecolor{screen}{RGB}{252,254,164}
\definecolor{applications}{RGB}{248, 223, 222}
\definecolor{keyboard}{RGB}{217, 231, 252}  
\definecolor{wifi}{RGB}{245, 183, 131}
\definecolor{calls}{RGB}{179,199,229} 
\definecolor{notifications}{RGB}{218,63,52}
\definecolor{battery}{RGB}{255,208,233}

\definecolor{customyellow}{HTML}{E7DECF} 
\definecolor{custompink}{HTML}{C27BA0}
\definecolor{customblue}{HTML}{94B5D7}   
\definecolor{customred}{HTML}{B50000}
\definecolor{customgreen}{HTML}{93c17d}

\maketitle

\begin{abstract}
Mental health conditions affect hundreds of millions globally, yet early detection remains limited. While large language models (LLMs) have shown promise in mental health applications, their size and computational demands hinder practical deployment. Small language models (SLMs) offer a lightweight alternative, but their use for social media–based mental health prediction remains largely underexplored. In this study, we introduce Menta, the first optimized SLM fine-tuned specifically for multi-task mental health prediction from social media data. Menta is jointly trained across six classification tasks using a LoRA-based framework, a cross-dataset strategy, and a balanced accuracy–oriented loss. Evaluated against nine state-of-the-art SLM baselines, Menta achieves an average improvement of 15.2\% across tasks covering depression, stress, and suicidality compared with the best-performing non–fine-tuned SLMs. It also achieves higher accuracy on depression and stress classification tasks compared to 13B-parameter LLMs, while being approximately 3.25$\times$ smaller. Moreover, we demonstrate real-time, on-device deployment of Menta on an iPhone 15 Pro Max, requiring only approximately 3GB RAM. Supported by a comprehensive benchmark against existing SLMs and LLMs, Menta highlights the potential for scalable, privacy-preserving mental health monitoring.
Code is available at: \url{https://hong-labs.github.io/menta-project/}
\end{abstract}

\section{Introduction}

Mental health disorders such as depression, anxiety, and suicidality affect hundreds of millions of people worldwide and constitute one of the leading contributors to the global burden of disease \citep{world2017depression}. Between 30\% and 50\% of people globally experience stress \citep{piao2024continuous}; an estimated 5.7\% of adults suffer from depression \citep{WHO2023depression}; and more than 720,000 people die by suicide each year \citep{WHO2025suicide}. As mental health issues have continued to rise globally over the past few decades \citep{goodwin2022trends,piao2024continuous,weaver2025global}, there is an urgent need to enhance our understanding, diagnosis, and monitoring of these conditions through advanced technologies.

Despite growing demand for mental health support, significant diagnosis gaps persist globally. Barriers such as limited clinic hours, geographic inaccessibility, and workforce shortages hinder timely assessment and early self-awareness \citep{WHOdiagnosis}. Structural, patient-level, and systemic obstacles further delay diagnosis for those with serious mental illness \citep{wiesepape2025behind}. This mismatch between high need and limited resources reflects the constraints of traditional diagnostic models \citep{stein2022psychiatric}. Moreover, these traditional delivery of in-person sessions and self-reported questionnaires struggles to scale, and faces systemic bottlenecks, including workforce shortages \citep{WHO_SIMH,brahmbhatt2024access}, high costs \citep{patel2018lancet}, geographic disparities \citep{yang2025wait}, and stigma \citep{clement2015impact}. Mental health wait times can range from 3 to 18 months, varying by region and provider availability \citep{mcmahan2022using,yang2025wait,patel2018lancet}. With the growing need for early detection and real-time monitoring, developing more seamless methods to identify early signs is critical for delivering timely support \citep{patel2018lancet}. This underscores the importance of portable, resource-efficient solutions such as small language models for scalable mental health detection and intervention.

\begin{figure}[t]
  \centering
  {
    \includegraphics[width=\textwidth, trim=0 95 0 20, clip]{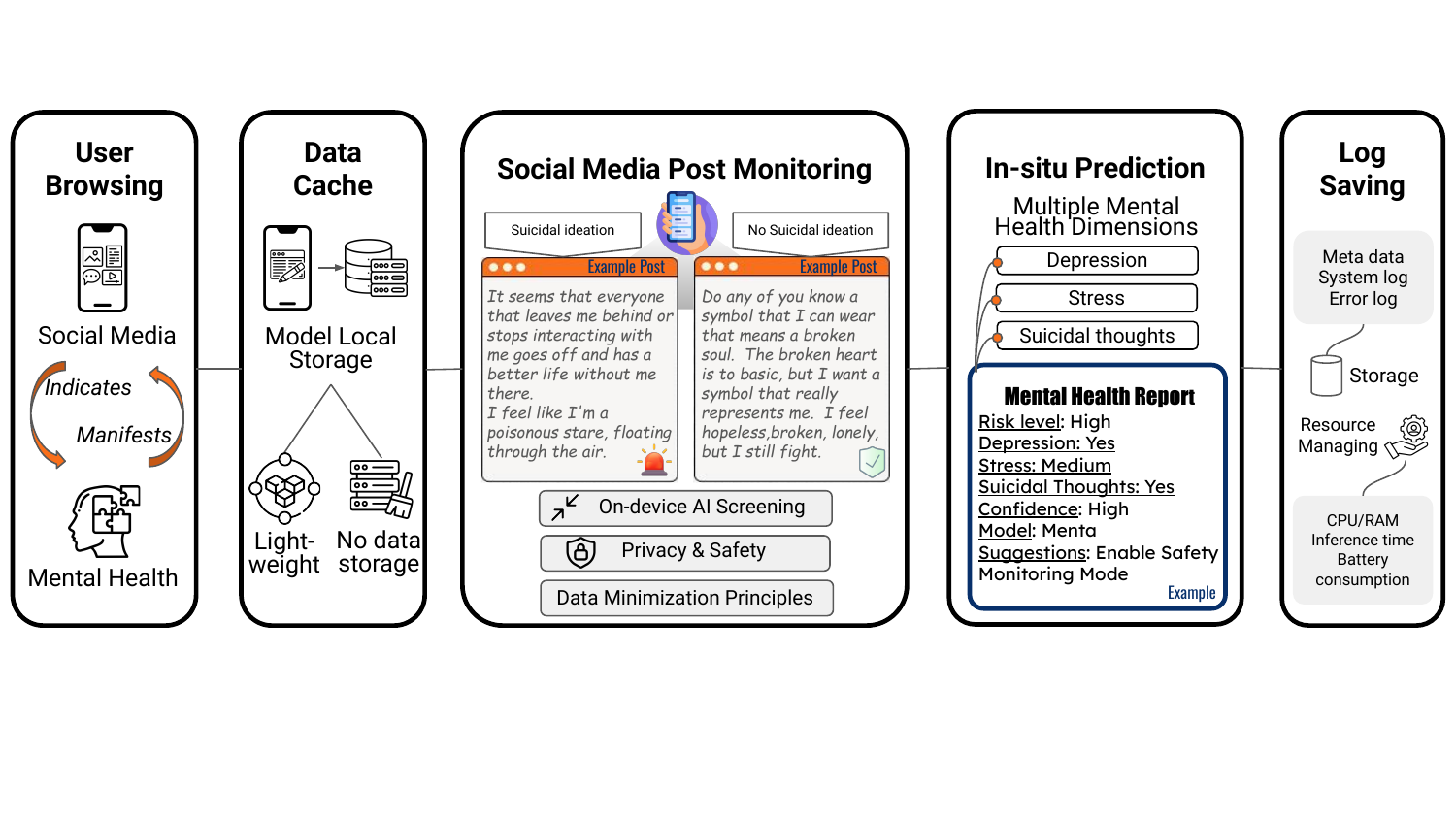}
    \label{workflow}
  }
  \caption{On-device mental health monitoring workflow using SLMs, from user browsing of social media, to data collection, post monitoring, in-situ prediction and meta log saving.}
  \label{fig:workflow}
\end{figure}


The rise of portable, on‑device technologies offers a promising path for the early detection of mental health issues, particularly through passive, real‑time analysis of social media posts and smartphone applications. Smartphone-based linguistic data from social media enables scalable, passive monitoring of psychological states, facilitating early identification of mood disturbances, stress, or suicide risk. For example, such systems can support suicide prevention efforts \citep{arean2016mobile} and capture patterns in social media use that are relevant to mental well-being \citep{hamilton2025improving}. Running locally on personal devices, these models preserve user privacy while analyzing language inputs for indicators of depression or stress \citep{shin2023fedtherapist}. As a result, on-device deployment enables scalable, privacy-preserving mental health monitoring, extending detection beyond clinical settings and making access easier for a broader population.

In this context, language models offer a powerful means of addressing the limitations of traditional approaches to mental health monitoring \citep{xu2024mental}. Language models outperform supervised learning classifiers in mental health prediction tasks for capturing deeper context, information acquisition, and generalization capabilities \citep{jin2025applications}, making them more practical for web-scale and real-world mental health monitoring.

However, the application of large language models (LLMs) faces significant challenges: they are computationally expensive to train and deploy, and they raise privacy concerns in terms of processing sensitive data on remote servers \citep{jin2025applications}. Their large size also makes them impractical for resource-constrained on-device settings, underscoring the need for more efficient alternatives. By contrast, Small Language Models (SLMs), offer comparable or even superior potential for digital health applications due to their lightweight architectures, lower inference costs, and feasibility of local deployment \citep{abdin2024phi,jia2025beyond}. Deploying on digital devices locally, a privacy-preserving, on-device workflow is expected for mental health prediction from social media activity using lightweight SLMs which emphasizes in-situ AI screening across multiple mental health dimensions without storing personal data, enabling real-time, low-resource, and ethically-aligned monitoring (Figure \ref{fig:workflow}). Yet, the potential of SLMs for mental health prediction remains largely underexplored.

Despite work on the development of SLMs, there is no agreement yet on a standardized definition of SLMs \citep{wang2024comprehensive}. Given the lightweight requirement for such real-time mental health status detection tasks on social media posts, we define SLMs to be models ranging from a minimum of 1B to strictly under 7B parameters are commonly considered `small' in contrast to full-scale LLMs (10B+) while those more than 7B to be large-scale small language models (LSSLMs) \citep{abdin2024phi,team2024qwen2}.

In this work, we explore the potential of SLMs in predicting digital mental health outcomes, and prove that SLMs, when carefully designed and optimized, serve as practical alternatives for multi-task mental health assessment on-device. To investigate this, we examine SLM performance across six mental health classification tasks spanning stress, depression, and suicidality. We introduce Menta, a compact model fine-tuned using a LoRA-based multi-task framework with cross-dataset training and a balanced accuracy–oriented loss. Menta performs strongly across diverse mental health tasks, particularly in stress and depression detection. We further demonstrate its practical value by deploying it on an iPhone 15 Pro Max, achieving real-time inference with minimal memory usage, enabling private and scalable on-device mental health support. Our key contributions are as follows:

\begin{itemize}
    \item We present Menta, a fine-tuned SLM tailored for multi-task mental health prediction from social media text, optimized for both accuracy and deployability.
    \item Menta leverages a weighted-loss, LoRA-based multi-task training framework designed to address data imbalance and maximize generalizability across multiple clinical classification tasks.
    \item We show that Menta offers a compelling balance between predictive accuracy and computational efficiency, outperforming several SLMs and rivaling LLMs on six mental health task.
    \item We provide an open-source deployment pipeline and model implementation for on-device inference, supporting privacy-preserving and resource-efficient mental health applications.
\end{itemize}

\section{Related Work}
\label{related_work}

\textbf{\emph{Social Media Posts: Indication of Mental Health. }}Social media posts offer a valuable lens into individuals’ emotions, thoughts, and mental health status. Globally, approximately 4.8 billion people, nearly 60\% of the population, use social media, spending a combined 11.5 billion hours on these platforms each day \cite{UMaine2023}. With the increasing use of social media platforms for emotional expression and peer support, user-generated text has become an important resource for understanding psychological well-being \citep{hussain2025optimised,zhunis2022emotion} and identifying early warning signs of mental distress \citep{chancellor2020methods,de2013predicting}. For instance, a study \citep{kim2020deep} built a classifier on Reddit posts to detect multiple mental disorders, while the EmoMent corpus \citep{atapattu2022emoment} shows that emotion annotations in user text correlate with mental illness indicators even in non-Western populations. These findings support the potential for social media–based models to serve as early warning systems for mental health support. 

Early approaches for textual information analysis among social media posts relied on traditional machine learning and feature-engineered models to classify or detect mental health status. For example, Jiang et al. used contextualized word embeddings such as ELMo and BERT to detect mental health conditions from Reddit posts \citep{jiang2020detection}, showing that deep contextualized models significantly outperform traditional bag-of-words or static embeddings for mental health classification tasks. Similarly, Sarkar et al. developed a multi-task learning model that jointly predicts depression and anxiety from Reddit posts, which outperforms single-task baselines by leveraging shared features across related conditions \citep{sarkar2022predicting}. Another studies developed a scalable deep-learning screening tool for suicide risk with potential clinical implications for early intervention such as CNN \citep{coppersmith2018natural}, LSTM \citep{coppersmith2018natural} and SVM \citep{ji2018supervised}.


More recently, LLMs have been adapted to this domain. A prior study used ChatGPT in zero‑shot classification tasks across three mental health domains (stress, depression, suicidality) on social media datasets, validating the ability of LLMs with no prior knowledge in mental health prediction tasks \citep{lamichhane2023evaluation}. Another study focuses only on the depression detection task from online text while collaborates the finding that fine-tuned LSSLMs achieve great improvements over prior state‑of‑art models \citep{shah2025advancing}. Moreover, the interpretability of LLMs in prediction tasks such as depressive symptoms \citep{bolegave2025gold,chen2025generating,belcastro2025detecting} and emotional states \citep{yang2023towards} have also been explored, with the finding that prompt engineering with emotional cues and few‑shot examples improves performance. Importantly, compared to traditional algorithmic models, lagnuage models offer not only strong predictive capabilities but also human-readable interpretations and the flexibility to support multiple downstream tasks within a unified framework.

A more comprehensive study introduced Mental-LLM \citep{xu2024mental}, demonstrated that instruction-tuned LLMs show large gains in balanced accuracy and can outperform task-specific baselines, with the best finetuned models (Mental‑Alpaca and Mental‑FLAN‑T5) outperform GPT‑3.5 by approximately 10.9\% and beat GPT‑4 by about 4.8\%. They also improved model generalizability fine-tuned on diverse datasets. Follow-up work has extended this line of research with domain-specialized models.  Similarly, Shi et al. proposed a LSSLM called MentalQLM for mental health tasks by using a base backbone (7B) and then applying a dual LoRA strategy \citep{hu2022lora}, achieving comparable or even superior performance to much larger LLMs on several mental health diagnostic tasks \citep{shi2025mentalqlm}. While these studies demonstrate the feasibility of using LLMs for mental health outcome prediction, significant challenges persist, including concerns over privacy and security \citep{sarwar2025fedmentalcare}, deployment limitations \citep{maurya2025exploring}, and high computational costs \citep{wang2023chatgpt}.

\textbf{\emph{Small Language Models and Mental Health.}} To address this gap of application bottlenecks, SLMs emerged as a pragmatic alternative of LLMs \citep{van2024survey}, optimized for efficiency through techniques such as quantization \citep{frantar2022gptq}, knowledge distillation \citep{kang2023knowledge} and light‑weight architectures \citep{wang2025survey}. SLMs were designed to deliver core Natural Language Processing capabilities such as text classification \citep{pecher2024comparing}, summarization \citep{wang2025distilling}, and question‑answering \citep{lee2024can}, in resource‑constrained settings including mobile devices, embedded systems, and edge computing. 

Offering a compelling alternative to LLMs, SLMs reduce computational overhead while retaining strong performance in many specialized tasks. Moreover, recent work shows that in classification settings with limited data, well‑tuned smaller models can match or even surpass larger models when the task is focused and domain‑specific including text classification \citep{lepagnol2024small,luo2023exploring} and healthcare domains \citep{gondara2025small}. These findings highlight that SLMs offer an efficient, scalable, and domain‑optimized solution, making them a promising direction for mental‑health detection from social media. However, previous work has typically focused on a single task or domain and has not explored multi-domain settings that combine multiple mental health conditions such as depression, stress, and suicidality.

Unlike prior LLM-based \citep{sarkar2022predicting,xu2024mental,yang2024mentallama,kim2024mhgpt,shi2025mentalqlm} which range from 7B to 70B parameters which are impractical for deplyment in resource-constrained or privacy-sensitive settings, lightweight SLMs emphasize deployability, reporting inference latency, memory footprint, and approximate cost efficiency. Prior work evaluated several SLMs against LLM baselines across multiple individual mental health understanding tasks with zero-shot and few-shot prompting trained on social media datasets, and concluded that few-shot prompting helps SLMs more \citep{jia2025beyond}. Also, Kim et al. developed a SLM called mhGPT, which outperforms larger models such as MentaLLaMA and Gemma in mental health tasks despite having far fewer parameters and using less data \citep{kim2024mhgpt}. A prior study also suggests SLMs suitable to be integrated in clinical workflows for structured and clinically meaningful tasks beyond generic detection \citep{aich2024using}. However, these studies lack systematic task balancing, rarely optimized with parameter-efficient fine-tuning (LoRA), meaning adaptation is either minimal (prompting) or heavy (full fine-tuning).

\textbf{\emph{Deployment and On-Device Language Models.}} On‑device AI offers a compelling opportunity for mental health monitoring by enabling personalized assessment in real time, while preserving user privacy, reducing latency, and expanding accessibility. Running inference directly on a user’s smartphone avoids sending sensitive linguistic and behavioural data to cloud servers, thereby lowering risks of data leakage and minimizing network dependencies \citep{mandal2025towards}. Furthermore, mobile and edge deployment largely improves accessibility by bringing mental health support tools directly to users, addressing traditional barriers such as geographic inaccessibility and long wait times for in‑clinic screening \citep{bunyi2021accessibility}. Additionally, because on‑device models eliminate round‑trip communication with remote servers, they deliver near‑instant responses and support real‑time detection of changes in language or behaviour indicative of stress or mood deterioration \citep{ni2025scoping}.

Mobile and edge-based mental health monitoring using smartphones and wearable sensors has gained momentum, offering significant advantages in latency, energy efficiency, and data privacy over cloud-dependent systems. Recent work has demonstrated effective deployment of SLMs on consumer devices, such as the Samsung Galaxy S24, for tasks like document assistance \citep{pham2024slimlm} and health prediction \citep{wang2025healthslm}, with substantial reductions in memory usage and inference latency. Conceptual analyses further argue that SLMs are better suited than LLMs for real-world, interactive applications due to their lightweight nature and ease of deployment under resource constraints \citep{belcak2025small}. On-device or near-device processing eliminates the need for continuous cloud communication, thereby enabling real-time detection while safeguarding user data—an essential consideration for privacy \citep{lu2024small}.

While recent studies have demonstrated the ability of SLMs on individual mental health tasks in zero- or few-shot settings, existing work remains limited in several key areas. Most notably, prior efforts have primarily focused on single-task classification, often failing to generalize well to more complex, multi-class, and imbalanced datasets. Moreover, little attention has been given to the real-world applicability of these models in terms of on-device deployment under resource constraints. To address these gaps, we propose Menta, a LoRA-based multi-task fine-tuning framework built on Qwen-3, incorporating weighted training to explicitly handle label imbalance. Our method is designed to achieve balanced and consistent performance across six diverse mental health prediction tasks, while maintaining the lightweight efficiency required for practical, privacy-preserving deployment on consumer devices.

\section{Method} \label{headings}

\subsection{Language Model Prompting} 

\subsubsection{Zero-shot Prompting}

\begin{figure}[t]
    \centering
    \includegraphics[width=\linewidth, trim=35 0 35 10, clip]{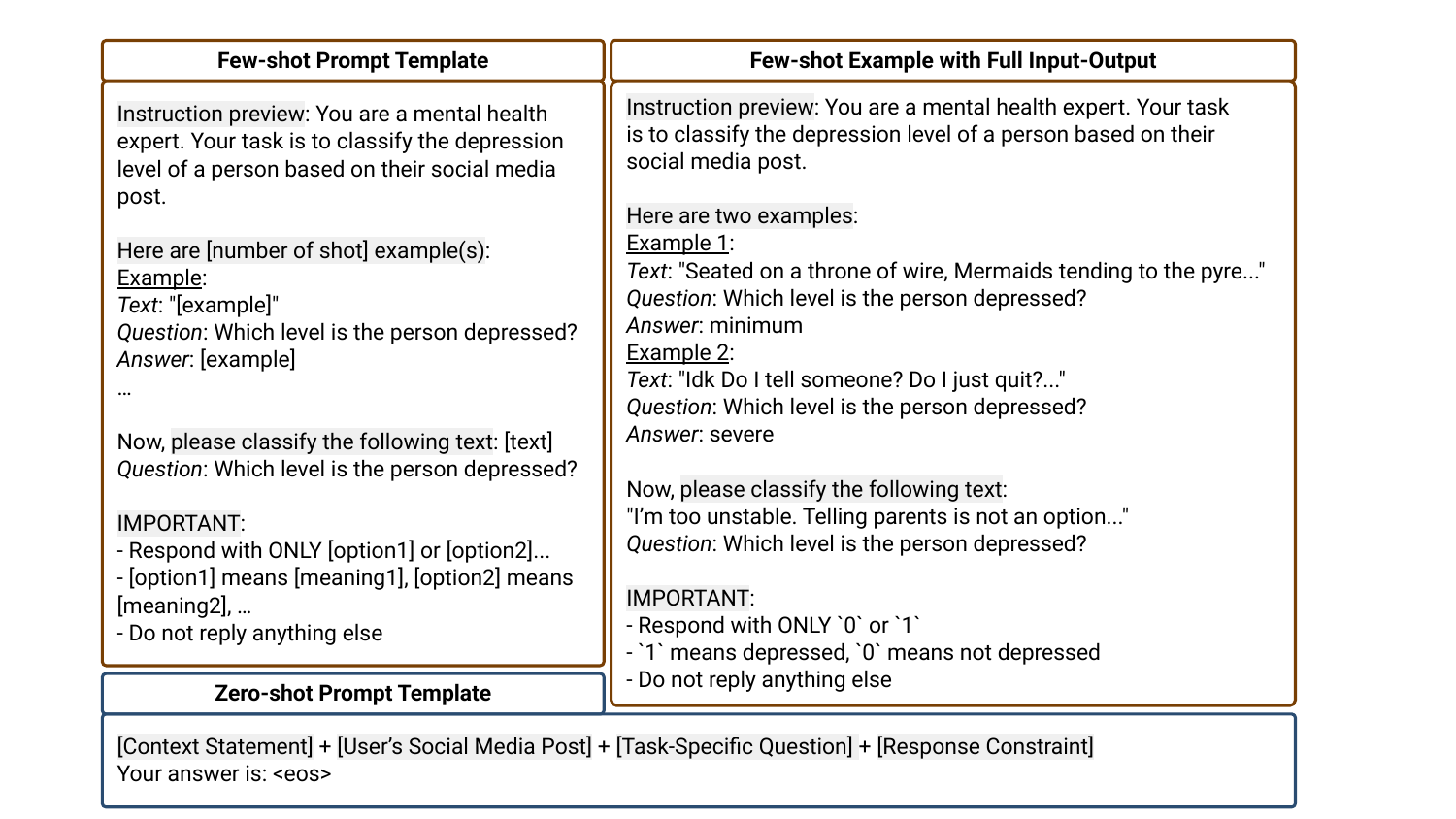}
    \caption{Prompt instruction templates for mental health classification tasks, specifically Zero-shot prompt, few-shot prompt template, and detailed few-shot prompt with examples.}
    \label{fig:prompt_templates}
\end{figure}

Zero-shot prompts were designed to reflect a domain-specific, psychologically grounded context without providing explicit task instructions or label options. Each prompt instructed the model to act as a psychologist evaluating a social media post for indicators of mental health conditions. The prompt construction followed a consistent structure composed of four key elements (Figure \ref{fig:prompt_templates}), designed to encourage the model to rely on contextual understanding of the user’s language and inferred psychological cues: 

\begin{enumerate}
    \item A context statement that framed the post as originating from social media and positioned the model in the role of a psychologist;
    \item The user’s text simulating a social media post;
    \item A task-specific question corresponding to one of the mental health classification tasks. 
    \item A response constraint explicitly requiring the model to respond only with a numeric label and avoid hallucination;
\end{enumerate}

\subsubsection{Prompt Formulation for Few-Shot and Fine-Tuned Models}

Both few-shot prompting and fine-tuning involve presenting SLMs with illustrative examples that demonstrate how how social media posts are categorized into relevant mental health categories. In both setups, the models are instructed to assume the role of a psychologist, evaluating each post based on the examples provided. The prompt is carefully designed to guide the model’s reasoning and classification behavior. Templates of the constructed prompt are shown in Figure~\ref{fig:prompt_templates}, including the following elements:

\begin{enumerate}
    \item A contextual statement framing the post as originating from social media and positioning the model in the role of a psychologist;
    \item Example social media posts paired with their corresponding mental health categories;
    \item A user text simulating a new social media post;
    \item A task-specific question aligned with one of the mental health classification objectives. 
    \item A response constraint explicitly requiring the model to respond only with a numeric label and avoid hallucination;
\end{enumerate}

\subsection{Model Fine-tuning} \label{sec:fine-tuning}

In this section, we focus on Qwen-3, aiming to develop a unified model that balances and enhances performance across distinct mental health tasks. Our tasks span six mental health prediction benchmarks, each formulated as a classification problem. We adopt Low-Rank Adaptation (LoRA) to fine-tune Qwen-3 (4B) for mental health prediction tasks. Instead of updating all model parameters, LoRA injects trainable low-rank matrices into specific weight projections, drastically reducing the number of trainable parameters while preserving reasoning capabilities. In our setting, LoRA adapters are applied to the query and value projections of the transformer attention layers, with rank $r=16$, scaling factor $\alpha_{\text{LoRA}}=32$, and dropout $0.05$. This configuration achieves efficient adaptation while keeping only $\sim0.1\%$ of the parameters trainable.

\begin{wrapfigure}{r}{0.45\textwidth}
    \centering
    \vspace{-5pt}
    \includegraphics[width=\linewidth, trim=240 195 135 75, clip]{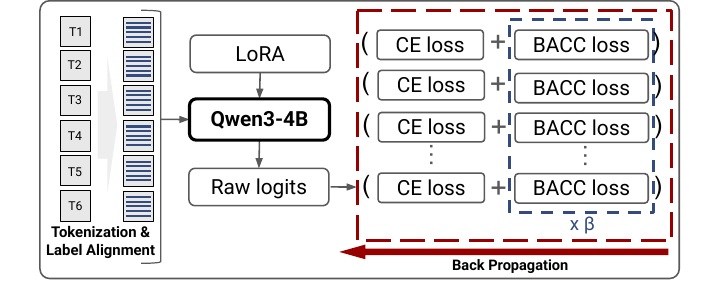}
    \caption{A multi-task training pipeline using Qwen-3 (4B) with LoRA, combining cross-entropy and balanced accuracy (BACC) losses weighted per task for joint optimization through backpropagation.}
    \label{fig:flowchart}
    \vspace{-5pt}
\end{wrapfigure}

We construct a unified multi-task dataset and apply weighted sampling across tasks to mitigate task imbalance. Specifically, the probability of sampling from task $t$ is proportional to a task weight $\lambda_t$, ensuring that underrepresented tasks are adequately seen during training. Within each task, class imbalance is addressed using inverse-frequency weights $w_c$ when computing the loss.

Traditional fine-tuning strategies typically rely on cross-entropy (CE) loss, which optimizes the log-likelihood of the true class. While effective, CE loss can bias the model toward majority classes, leading to degraded performance on minority classes. To address this, we introduce a novel \textit{balanced accuracy (BACC) surrogate loss} that provides a differentiable approximation of balanced accuracy, thereby directly encouraging the model to perform more evenly across classes, as shown in Figure \ref{fig:flowchart}.

For a given task $t$, the standard cross-entropy loss is defined as:
\begin{equation}
\mathcal{L}_{\text{CE}}^{(t)} = -\frac{1}{N_t} \sum_{i=1}^{N_t} \log p(y_i \mid x_i),
\end{equation}
where $N_t$ is the number of samples for task $t$, and $p(y_i \mid x_i)$ is the predicted probability for the true label $y_i$, obtained via the softmax over model logits.

To approximate balanced accuracy (BACC), we first compute a \textit{margin} for each class $c$:
\begin{equation}
m_{i,c} = z_{i,c} - \log \left( \sum_{k \neq c} \exp(z_{i,k}) \right),
\end{equation}
where $z_{i,c}$ denotes the model logit of sample $i$ for class $c$. This margin quantifies the relative confidence of the model in predicting class $c$ compared to all other classes.

We then apply a sigmoid function with sharpness parameter $\alpha$ to obtain a soft correctness score:
\begin{equation}
s_{i,c} = \sigma(\alpha \cdot m_{i,c}),
\end{equation}
where $\sigma(\cdot)$ is the sigmoid function.

Using these soft scores, the true positive rate (TPR) for each class $c$ is estimated as:
\begin{equation}
\widehat{\mathrm{TPR}}_c = \frac{1}{|I_c|} \sum_{i \in I_c} s_{i,c},
\end{equation}
where $I_c$ is the set of indices such that $y_i = c$. If no samples belong to class $c$, the corresponding $\widehat{\mathrm{TPR}}_c$ is defined as $0$.

The surrogate BACC loss is then defined as:
\begin{equation}
\mathcal{L}_{\text{BACC}}^{(t)} = 1 - \frac{1}{\sum_c \gamma_c} \sum_c \gamma_c \cdot \widehat{\mathrm{TPR}}_c,
\end{equation}
where $\gamma_c$ is a class scaling factor that adjusts the relative importance of each class.

The total multi-task loss is computed as a weighted combination of the cross-entropy loss and the surrogate BACC loss:
\begin{equation}
\mathcal{L}_{\text{total}} = \sum_{t} \lambda_t \left( \mathcal{L}_{\text{CE}}^{(t)} + \beta \cdot \mathcal{L}_{\text{BACC}}^{(t)} \right),
\end{equation}
where $\lambda_t$ is the task-specific weight and $\beta$ is a fixed hyperparameter that controls the trade-off between cross-entropy and BACC optimization. In this formulation:
\begin{itemize}
    \item $\lambda_t$: task-specific weight (externally specified),
    \item $\beta$: fixed trade-off parameter between $\mathcal{L}_{\text{CE}}$ and $\mathcal{L}_{\text{BACC}}$,
    \item $\gamma_c$: class scaling factor,
    \item $\alpha$: sigmoid sharpness parameter controlling sensitivity to the margin.
\end{itemize}

This approach encourages fairness across tasks and classes by directly optimizing a smooth surrogate of balanced accuracy. Unlike standard CE loss, it explicitly penalizes performance gaps between majority and minority classes, improving consistency in multi-task fine-tuning. An ablation study was conducted to compare different training strategies. Specifically, we evaluated the impact of single-task versus multi-task fine-tuning approaches for mental health prediction.

\subsection{On-Device Deployment}
We developed an on-device mental health evaluation system for iOS using llama.cpp\footnote{https://github.com/ggml-org/llama.cpp} as the inference backend, leveraging GGUF V3–quantized models optimized for mobile hardware. The models were exported from PyTorch to the GGUF format via the transformers-to-gguf conversion pipeline, enabling 4-bit quantization (Q4\_K\_M) for efficient CPU and GPU execution. The app was implemented in Swift/SwiftUI and integrates llama.cpp through a C++ bridge layer, allowing direct inference within the iOS runtime without external servers.

At runtime, inference is accelerated using Apple Metal for matrix operations, with automatic CPU fallback when GPU utilization is saturated. Thread-level parallelism (up to 8 threads) is managed through llama.cpp’s built-in thread scheduling API, ensuring optimal performance across Apple’s efficiency and performance cores. Model weights are memory-mapped to minimize RAM usage and enable lazy loading of tensors.

For stability, the system uses a 4,096-token context window with efficient KV-cache management and batched inference to avoid fragmentation during long input processing. Long social media posts are truncated to 6,000 characters. We deployed and evaluated three models locally, including our fine-tuned Menta (2.33 GB), Phi-4 Mini (2.40 GB), and Qwen-3 (2.30GB), on an iPhone 15 Pro Max (A17 Pro chip, 8GB RAM) with open-sourced code\footnote{https://anonymous.4open.science/r/Menta-6CAF}. The deployment framework also supports other quantization families (Q2\_K–Q8\_0), allowing trade-offs between latency and accuracy.

\section{Experiments} \label{others}

\subsection{Datasets and Task Definitions} \label{sec:method_dataset}

For our fine-tuning model Menta, we adopted four high-quality corpora for task-specific detection of depression, stress, and suicidal ideation collected from the social media platform Reddit, where disorder-related samples were excluded from the texts and the remaining data were annotated by domain experts. The datasets used in this study include the following:

\newcolumntype{L}[1]{>{\raggedright\arraybackslash}p{#1}}

\textbf{Dreaddit} \cite{turcan2019dreaddit}: From a multi-domain stress corpus, 3,500 segments were manually annotated via Amazon Mechanical Turk as either “stress” or “not stress”, ensuring coverage across diverse domains. 

\textbf{DepSeverity} \cite{naseem2022early}: This dataset is aimed at accurately identifying users' depression severity levels, annotated by CLEF eRisk organizers based on DSAS and clinical standards, categorizing the dataset into four severity levels: Minimal, Mild, Moderate, and Severe. 

\textbf{SDCNL} \cite{haque2021deep}: This dataset of 1,895 Reddit posts was collected from communities related to depression and suicidal ideation. Posts from suicide-related communities were labeled as ``suicidal ideation,'' and posts from depression communities were labeled as "non-suicidal ideation.", subsequently employed unsupervised labeling methods to correct potential misclassifications. 

\textbf{CSSRS-Suicide} \cite{gaur2019knowledge}: This dataset contains samples related to depressive suicidal ideation and behaviors. Five hundred users were randomly selected, and domain experts annotated the data using the Columbia-Suicide Severity Rating Scale \citep{brown2020c}. The resulting labels fall into five categories: supportive, indicator, ideation, behavior, and attempt. 

\begin{figure}[t]
    \centering
    \includegraphics[width = \linewidth]{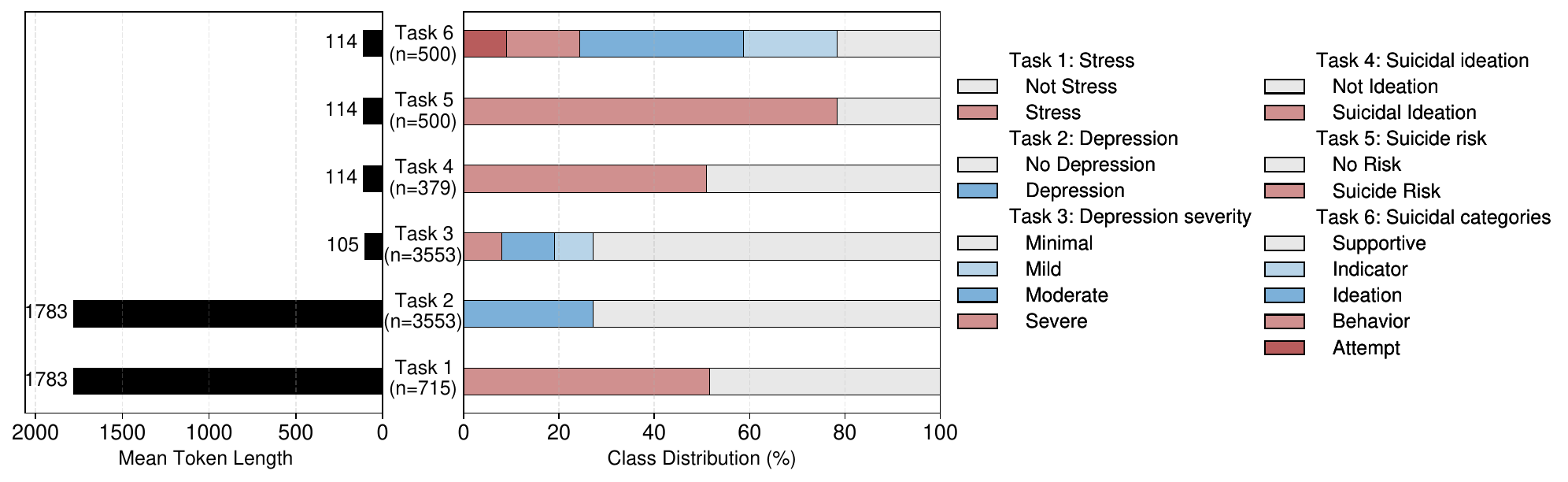}
    \caption{Class and token length distributions across six mental health classification tasks, highlighting label imbalance and diversity in annotation types.}
    \label{fig:class_distribution}
\end{figure}

As shown in Figure \ref{fig:class_distribution}, the Dreaddit dataset is used for Task 1 (binary stress classification), a subset of 715 posts was used, with 48.4\% labeled as stressful with an average post length of 114 $\pm$ 42 tokens. We utilized the DepSeverity dataset for Tasks 2 and 3, which supports two post-level binary classification in Task 2 and four-level severity of depression tasks in Task 3). Both Task 2 and 3 have 3,553 posts with an average length of 114 $\pm$ 43 tokens. Specifically, the class distribution for Task 3 is: 72.8\% Minimal, 8.2\% Mild, 11.1\% Moderate, and 7.9\% Severe. 

The SDCNL dataset addresses binary suicide ideation classification (labeled [1/0]) in Task 4 on 379 posts (49.1\% flagged), with average text length of 105 $\pm$ 12 tokens. The CSSRS-Suicide dataset provides user-level data for binary (Task 5) and five-level (Task 6) suicide risk classification, with 500 users and an average of 1,783 $\pm$ 2,178 tokens per user. For Task 6, the class distribution is: 21.6\% Supportive, 19.8\% Indirect, 34.2\% Ideation, 15.4\% Behavior, and 9.0\% Attempt. The levels of severity and classifications for the six mental health tasks are as follows:

\begin{itemize}
    \item Task 1: Not stressed [0]; Stressed [1].
    
    \item Task 2: No Depression [0] = Minimal; Depression [1] = Mild + Moderate + Severe.
    
    \item Task 3: Minimal [0]; Mild [1]; Moderate [2]; Severe [3].
    
    \item Task 4: No suicidal ideation [0]; Suicidal ideation [1].

    \item Task 5: No indicator of suicide risk [0] = Supportive; Indicator of suicide risk [1] = Indicator + Ideation + Behavior + Attempt. 

    \item Task 6: Supportive [1] = Emotional support but no risk signals; Indicator [2] = Indirect signs of vulnerability or concern; Ideation [3] = Explicit suicidal thoughts without action; Behavior [4] = Suicide-related behaviors short of attempts; Attempt [5] = Evidence of actual suicide attempts.
\end{itemize}

For data pre-processing, raw datasets were first loaded from task-specific CSV files, after which categorical labels were mapped to standardized numeric formats, with any unmappable entries removed. To preserve class distributions, stratified splitting was applied, using a 72/8/20 ratio for training, validation, and testing. 

\subsection{Baseline Experiments} 

For each model and classification task, we conducted experiments under five distinct prompting configurations: zero-shot, one-shot, two-shot, three-shot, and four-shot learning. Each configuration was evaluated independently, and for robustness, we performed five separate runs per setting. This setup enables a systematic comparison of performance across different levels of in-context learning while maintaining consistency across tasks and models.

\subsection{Small vs. Large Language Models}
In this study, we evaluate a diverse set of SLMs limited to under 7B parameters. \textbf{Phi-3 (3.8B)} and \textbf{Phi-3.5 (3.8B)} are lightweight Microsoft models designed for reasoning and efficiency~\citep{abdin2024phi}. \textbf{LLaMA 3.2 (3B)} is Meta’s compact release optimized for low-resource deployment~\citep{dubey2024llama}. \textbf{Gemma-3 (1B and 4B)} are Google’s instruction-tuned models emphasizing accessibility and safety~\citep{team2025gemma}. \textbf{Qwen-2.5 (3B)} and \textbf{Qwen-3 (4B)} are Alibaba’s models tailored for multilingual and domain adaptability~\citep{yang2025qwen3}. \textbf{Phi-4 Mini (3.8B)} extends Microsoft’s efficiency line~\citep{abouelenin2025phi}. \textbf{TinyLLaMA (1.1B)}~\citep{zhang2024tinyllama} and \textbf{Falcon (1.3B)}~\citep{almazrouei2023falcon} are streamlined open-source transformer models for edge deployment. \textbf{StableLM (3B)} is Stability AI’s open family prioritizing lightweight deployment and transparency~\citep{pinnaparaju2024stable}.

We developed a multi-task fine-tuned model called Menta employed a shared Qwen3-4B backbone with LoRA adapters, jointly trained on all six tasks. Training was balanced using task-specific sampling weights, and the proposed combined loss function (\S\ref{sec:fine-tuning}) was applied to each task. We compare the developed Menta model using the adjusted loss function with other selected fine-tuned SLMs, including Phi-4 Mini (3.8B), StableLM (3B) and Falcon-1.3B. In addition, we compare the SLMs against larger-scale baselines, Mental-Alpaca and Mental-FLAN-T5 \cite{xu2024mental}, developed by Xu et al. through multi-task instruction fine-tuning of Alpaca (7B) \cite{taori2023stanford} and FLAN-T5 (11B) \cite{chung2024scaling} using the same datasets and tasks. 

\subsection{Dependant Model vs. General Model}
To provide a comprehensive evaluation, we conducted an ablation study comparing single-task and multi-task fine-tuning strategies. In the single-task fine-tuning setting, we trained six independent models, each corresponding to one of the mental health prediction tasks. Each model used Qwen3-4B as the backbone with LoRA adapters applied to enable efficient parameter updates. This setting represents a task-specialized approach, where each model is optimized exclusively for its own objective without considering cross-task transfer. This ablation disentangles the effects of multi-task training from the BACC surrogate loss, clarifying the design choice’s contribution.

\subsection{Evaluation Metrics}
We evaluated model performance using accuracy (ACC) as the primary metric for both binary and multi-class classification tasks, selected for interpretability and broad applicability across tasks with differing label granularities and defined as:

\[
\text{Accuracy} = \frac{TP + TN}{TP + TN + FP + FN}
\]

where $TP$, $TN$, $FP$, and $FN$ denote true positives, true negatives, false positives, and false negatives, respectively.

To address label imbalance, we additionally reported balanced accuracy (BACC), which averages true positive rates across classes:

\[
\text{BACC} = \frac{1}{C} \sum_{c=1}^C \frac{TP_c}{TP_c + FN_c}
\]

where $C$ is the total number of classes, and $TP_c$, $FN_c$ are true positives and false negatives for class $c$. This metric is included to enable a fairer comparison among SLMs. Notably, prior work has not evaluated this metric for LLMs, and such evaluation falls outside the scope of this study.

To demonstrate the feasibility, usability, and lightweight nature of the SLMs for real-time, privacy-preserving mental health inference directly on mobile devices, the following metrics are evaluated for the deployment:
\begin{itemize}
\item Time to First Token  (TTFT, sec): The measurement of how long it takes for the model to generate the first token in response to a prompt (response latency).

\item Input Token Processing Speed (ITPS, tokens/sec): An indication of the rate at which input tokens are handled by the model (input throughput).

\item Output Token Processing Speed (OTPS, tokens/sec): A representation of how quickly the model generates output tokens after beginning generation (output efficiency).

\item Output Elapsed Time (OET, sec): The duration from the start of output generation to the end.

\item End-to-End Latency (Total Time): Total elapsed time from prompt submission to final response (overall system efficiency).

\item Memory Consumption (RAM, GB): Amount of system memory used by the model during inference.
\end{itemize}

\section{Results}

We present the results of the experiments for zero-shot and few-shot learning baselines in \S\ref{sec:zero-shot-learning}, analyze the outcomes of fine-tuning tasks in \S\ref{sec:fine-tuning}, evaluate deployment considerations in \S\ref{sec:deployment}, and demonstrate specific cases in \S\ref{sec:case-study}.

\begin{figure}[t]
  \centering
  {
    \includegraphics[width=\textwidth]{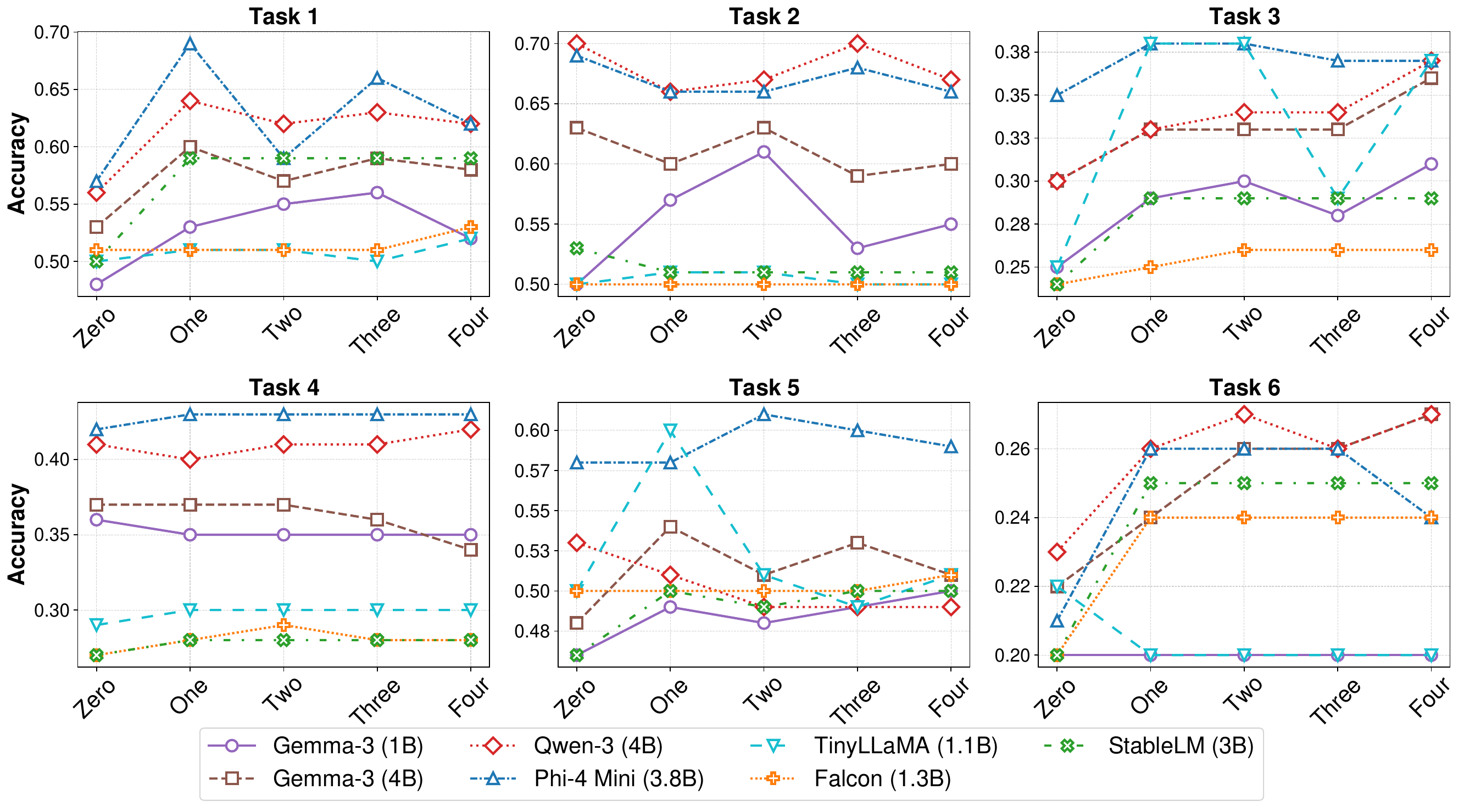}
  }
  \vspace{-1.5em}
  \caption{Performance of various SLMs across mental health tasks under zero-shot and few-shot settings. The x-axis indicates the number of shots used.}
  \label{fig:fewshot_results}
\end{figure}

\subsection{SLM Baselines} \label{sec:zero-shot-learning}

Evaluating the zero-shot and few-shot learning performance for nine SLMs, our results (Table \ref{tab:fewshot_results}) indicate that most SLMs possess only a basic capability to accurately classify social media posts related to depression, stress, and suicidal ideation without fine-tuning. Two models, Qwen-3 (4B) and Phi-4 Mini (3.8B), demonstrated the strongest performance, achieving average zero-shot accuracies of 45.5\% and 47.0\% across the six tasks respectively. Among the models, increasing the number of examples generally enhances performance despite exhibiting some variability across the six tasks, with the most notable improvement observed in Qwen-3 (4B) and Phi-4 Mini (3.8B), as shown in Figure \ref{fig:fewshot_results}. Despite modest improvements with few-shot examples, the overall accuracy remains insufficient for practical, real-world applications, which requires the fine-tuning models for further improvement.

Upon examining the model outputs, we observed that SLMs with approximately 1B parameters generally exhibit limited task comprehension and often fail to produce responses in the required format (e.g., binary outputs such as 1 for depression and 0 for no depression). Although their tendency to default to standard responses may result in seemingly acceptable evaluation metrics, these models fundamentally lack the robustness required for reliable real-world deployment.

\begin{figure}[t]
  \centering
  \begin{minipage}{0.48\textwidth}
    \centering
    
    \includegraphics[width=\textwidth]{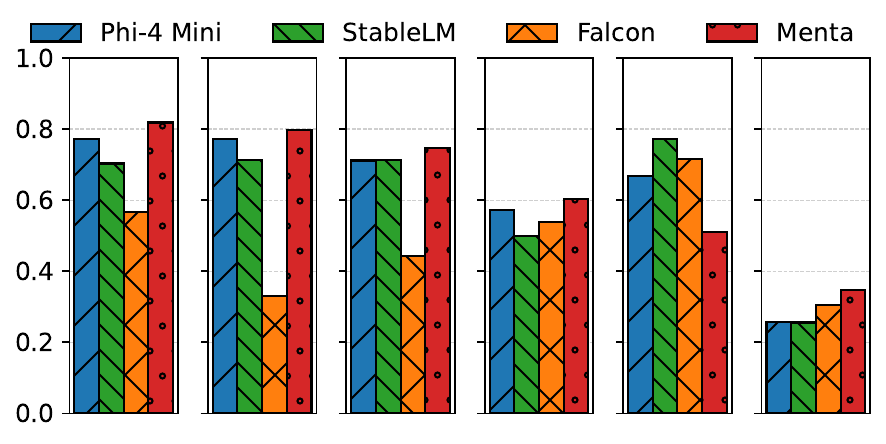} \
    (a) ACC Comparison\\[-1ex]
  \end{minipage}
  \hfill
  \begin{minipage}{0.48\textwidth}
    \centering
    
    \includegraphics[width=\textwidth]{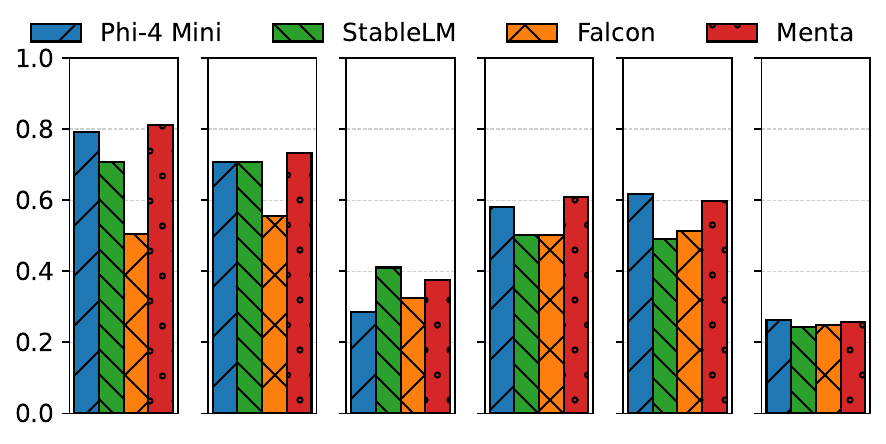} \
    (b) BACC Comparison\\[-1ex]
  \end{minipage}
  \vspace{-1em}
  \caption{Accuracy (ACC) and Balanced Accuracy (BACC) scores for six tasks (left to right), comparing models Phi-4 Mini (blue), StableLM (green), Falcon (orange), and Menta (red), with Menta consistently achieving the highest performance.}
  \label{fig:acc_bar_chart}
\end{figure}

\subsection{Fine-tuning Tasks} 

We demonstrate that the general Menta model consistently outperforms both zero-shot and few-shot learning settings, achieving an average improvement of 15.2\% across all six tasks over Qwen-3. Notably, it achieves the largest improvements in Task 3, Task 1, Task 6 and Task 2 over the zero-shot setting, with gains of 44.4\%, 25.4\%, 11.5\% and 10.2\%, respectively. These results suggest that, compared to zero-shot and few-shot paradigms, fine-tuning with domain-specific mental health data enables SLMs to perform more effectively and reliably on these tasks.

\subsubsection{Fine-tuned SLMs}

By fine-tuning various SLMs, our results show that across six mental health prediction tasks, our model Menta achieves the highest average performance (ACC=0.637), outperforming Phi-4 Mini by 1.2\%, StableLM by 2.8\%, and Falcon by nearly 15.4\%, demonstrating its superior accuracy and robustness across diverse mental health tasks, as shown in Figure \ref{fig:acc_bar_chart}(a). Our results also demonstrate that incorporating the adjusted loss function led to improvements in BACC across all trained SLMs, Menta achieves the highest BACC (0.564) and outperforming Phi-4 Mini by 1.9\%, StableLM by 3.1\%, and Falcon by nearly 17.0\%, as shown in Figure \ref{fig:acc_bar_chart}(b). These results demonstrate that Menta not only leads in overall accuracy but also offers more reliable performance across diverse task settings.

\subsubsection{Dependant Model vs. General Model}

\begin{figure}[t]
  \centering
  \subfloat[ACC Comparison]{%
    \includegraphics[width=0.47\textwidth]{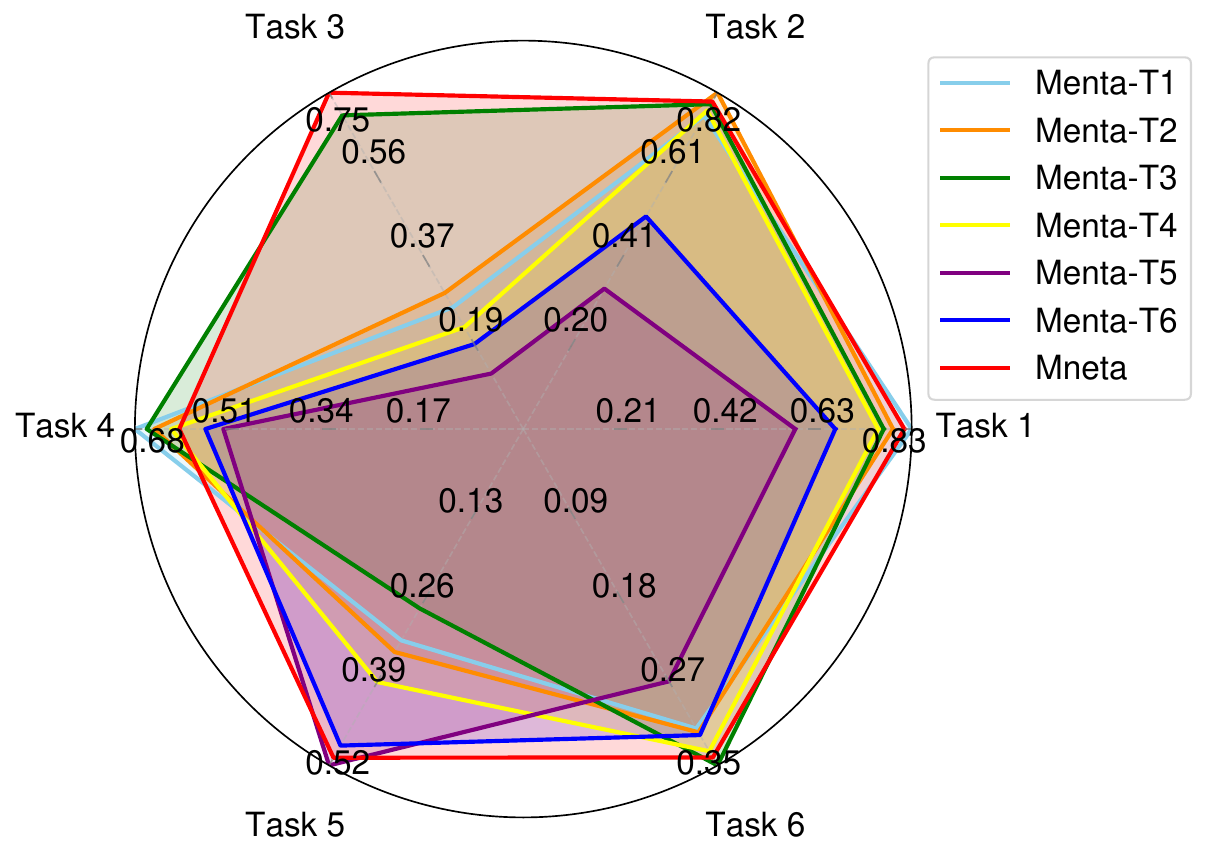}%
  }
  \subfloat[BACC Comparison]{%
    \includegraphics[width=0.47\textwidth]{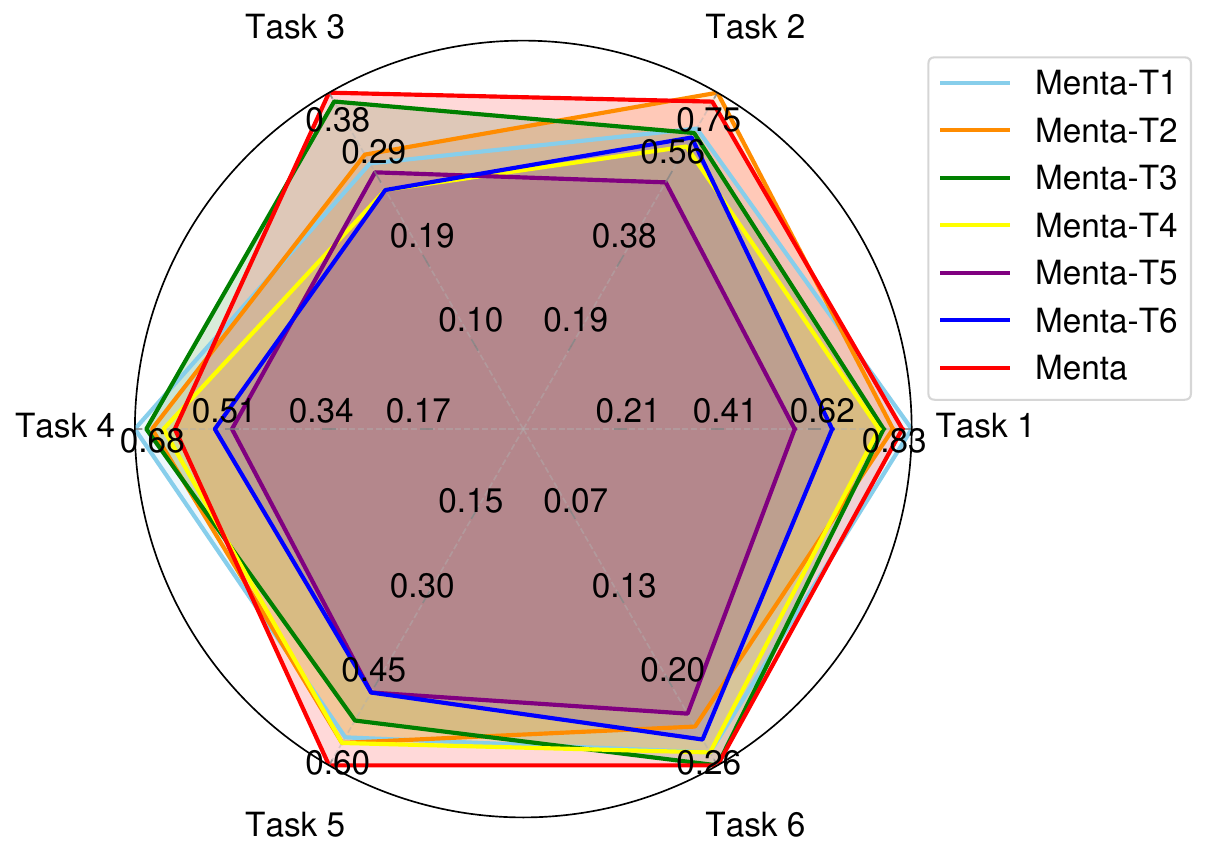}%
  }
  \caption{Model performance across six tasks for the individually fine-tuned models Menta-T1 to Menta-T6, compared with the general Menta model optimized using BACC-aware settings.}
  \label{fig:individual_models}
\end{figure}

We further compared single-task fine-tuning Menta against the general Menta trained on cross-datasets and we show that the general Menta model consistently outperformed models trained on individual tasks (Table \ref{tab:task_all_results}), suggesting robustness across varied mental health classification tasks and no multiple models needed for saving memory. Menta also exhibits lower standard deviations, underscoring its stability and consistency. We refer to the single-task variants as Menta-T1, Menta-T2, etc. As shown in Figure \ref{fig:individual_models}, while Menta-T1 and Menta-T2 perform strongly in tasks 1 and 2 respectively, they lack broader consistency across tasks. This ablation study results suggest that although single-task SLMs serve as effective specialists, Menta demonstrates superior versatility as a generalist, making it more practical and impactful for comprehensive mental health monitoring, particularly in mobile or edge deployment scenarios.

\begin{table}[t]
\centering
\small
\caption{Performance of Menta in comparison with trained LLMs Across Tasks, with the best performance highlighted in bold and the second-best underlined.} \label{tab:llm_slm}
\begin{tabular}{lcccccc}
\toprule
Model & Task 1 & Task 2 & Task 3 & Task 4 & Task 5 & Task 6 \\
\midrule
Menta (4B) [Ours] & \textbf{0.82} ± 0.01 & \textbf{0.80} ± 0.04 & \textbf{0.75} ± 0.01 & 0.60 ± 0.02 & 0.51 ± 0.02 & 0.35 ± 0.06 \\
Mental-Alpaca (13B) & \textbf{0.82} ± 0.01 & \underline{0.78} ± 0.01 & \textbf{0.75} ± 0.01 & \textbf{0.72} ± 0.00 & \underline{0.73} ± 0.05 & \underline{0.40} ± 0.03 \\
Mental-FLAN-T5 (13B) & \underline{0.80} ± 0.00 & 0.76 ± 0.00 & \underline{0.76} ± 0.00 & \underline{0.68} ± 0.01 & \textbf{0.87} ± 0.01 & \textbf{0.48} ± 0.01 \\
\bottomrule
\end{tabular}
\vspace{-1em}
\end{table}

To further understand model behavior across tasks, we visualized training and validation loss curves over epochs for each of the six tasks (Figure \ref{fig:loss}). The plots reveal that most tasks exhibit optimization with a consistent decrease in training loss. However, the validation loss curves are flatter or plateau earlier, suggesting limited generalization. This trend highlights the constrained capacity of SLMs to transfer learned representations across diverse mental health tasks. Notably, the training loss drops most sharply within the first few hundred samples, after which improvements become incremental. This suggests that the model rapidly captures the core patterns of the task early in training, and additional data primarily contributes to fine-tuning and refinement rather than introducing fundamentally new information. The smooth decline in the general Menta model’s training loss, coupled with its balanced task-wise performance as reflected in BACC, indicates that the model achieves stable and generalized learning rather than overfitting to individual tasks.

\subsubsection{Menta vs. LLMs}

Comparing the performance of Menta and fine-tuned LLMs in previous studies, Table \ref{tab:llm_slm} shows that the Menta model is superior and comparable with fine-tuned LLMs with 81.8\%, 79.8\% and 75\% accuracy for binary stress (Task 1) and binary depression (Task 2) classification and four-level depression (Task 3) respectively. However, for the other three tasks, Menta underperforms with an average of 13\% and 19\% compared to the Mental-Alpaca and Mental-FLAN-T5 mdoel. This gap emphasizes the ability of SLMs is not as good as LLMs for suicidal risk detection tasks.

Nonetheless, we found a trade-off between the size of the models and their performance. Although larger models (13B) achieve higher absolute performance across certain tasks, their performance improvements are not proportional to their parameter increase. When normalized by model size, the 4B Menta model delivers approximately three times higher performance-per-parameter compared to 13B models. Compared to 13B-parameter models, Mental-Alpaca and Mental-FLAN-T5, Menta is approximately 3.25$\times$ smaller, while delivering comparable or superior performance on Tasks 1 and 2, and maintaining competitive results across other tasks. This indicates that smaller models offer a more favorable trade-off between computational efficiency and accuracy, particularly in deployment-constrained mental health applications.


\begin{figure}[t]
  \centering
  \includegraphics[width=0.3\textwidth]{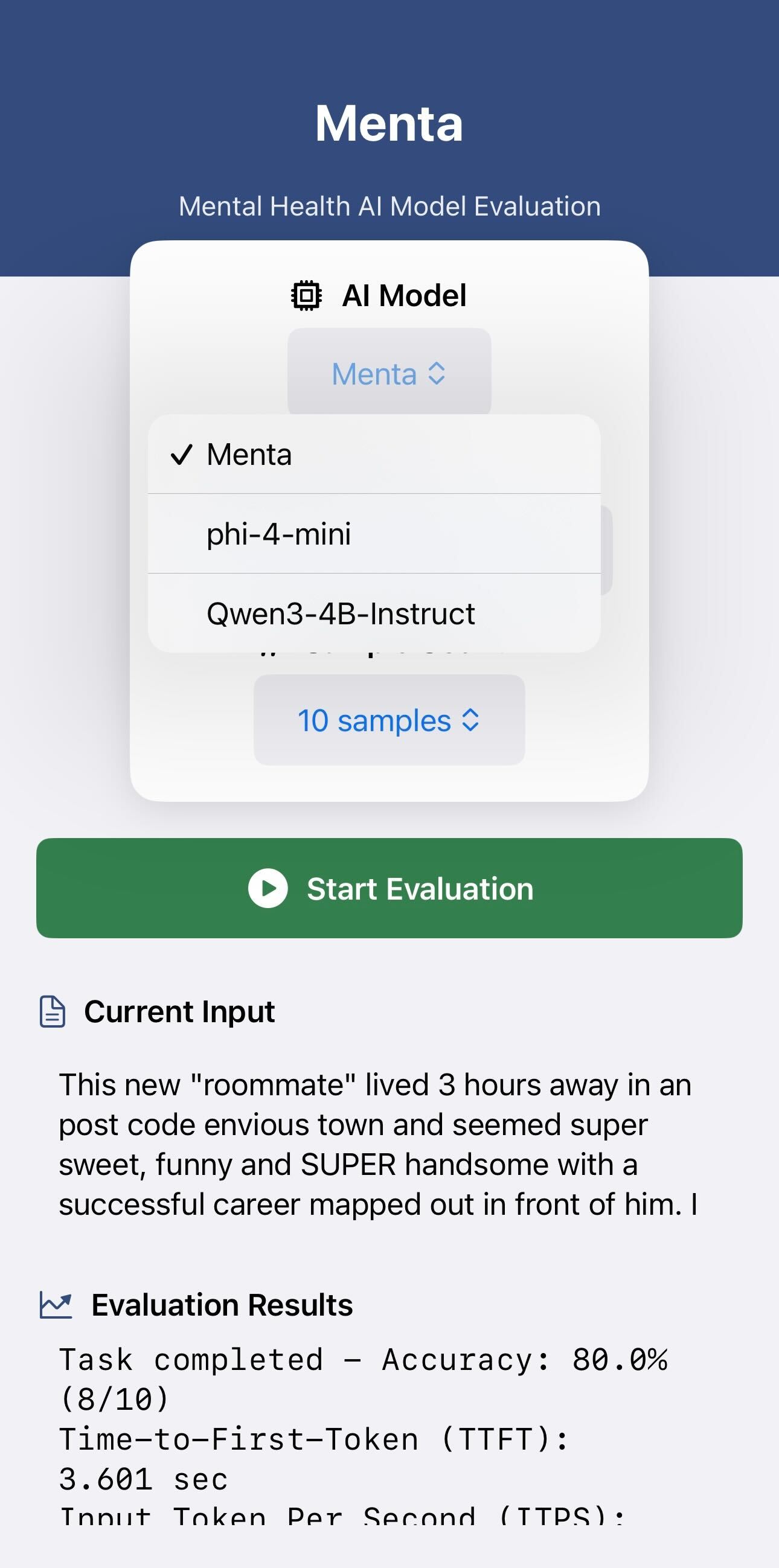}
  \hspace{0.02\textwidth}
  \includegraphics[width=0.3\textwidth]{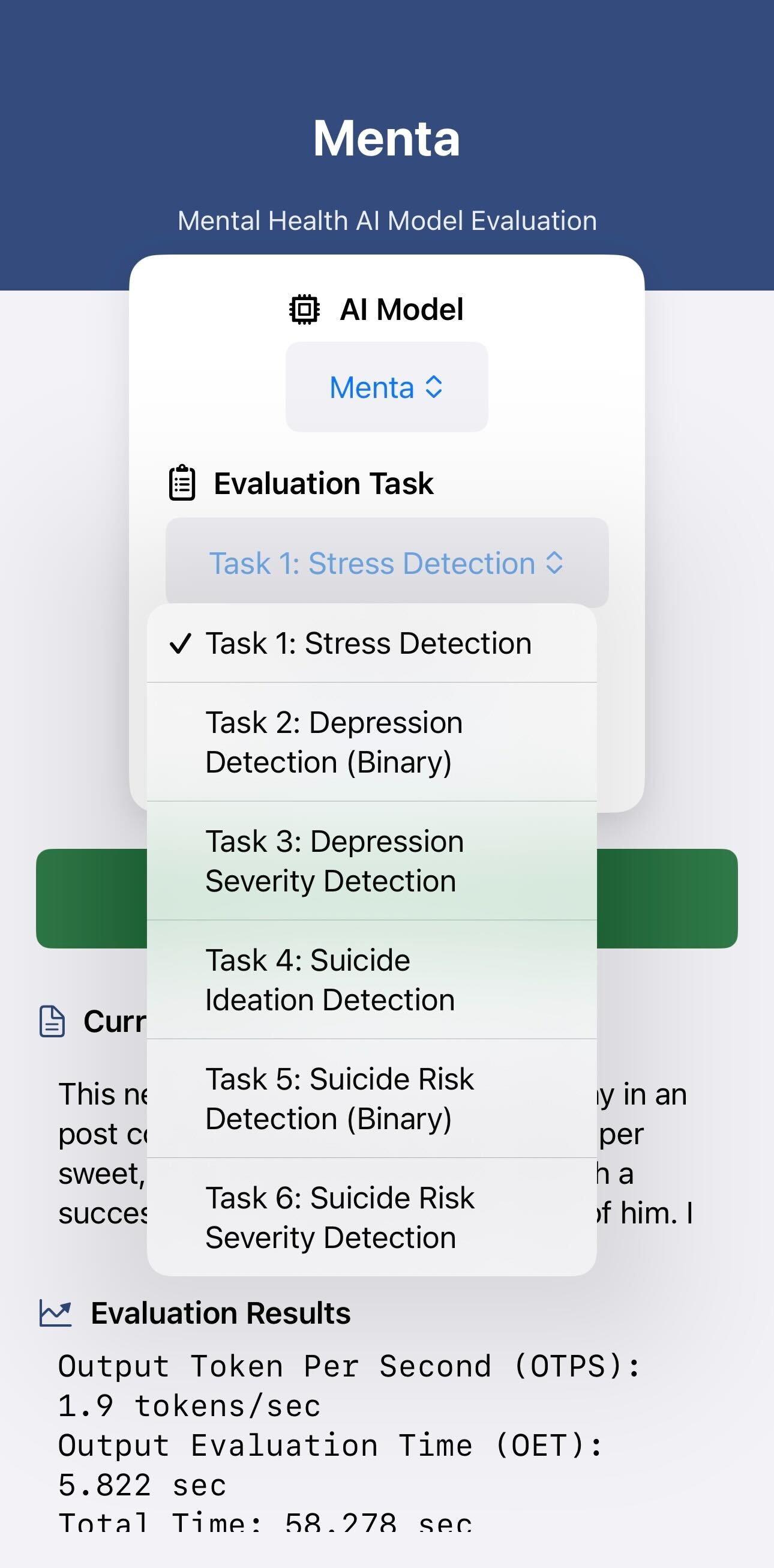}
  \hspace{0.02\textwidth}
  \includegraphics[width=0.3\textwidth]{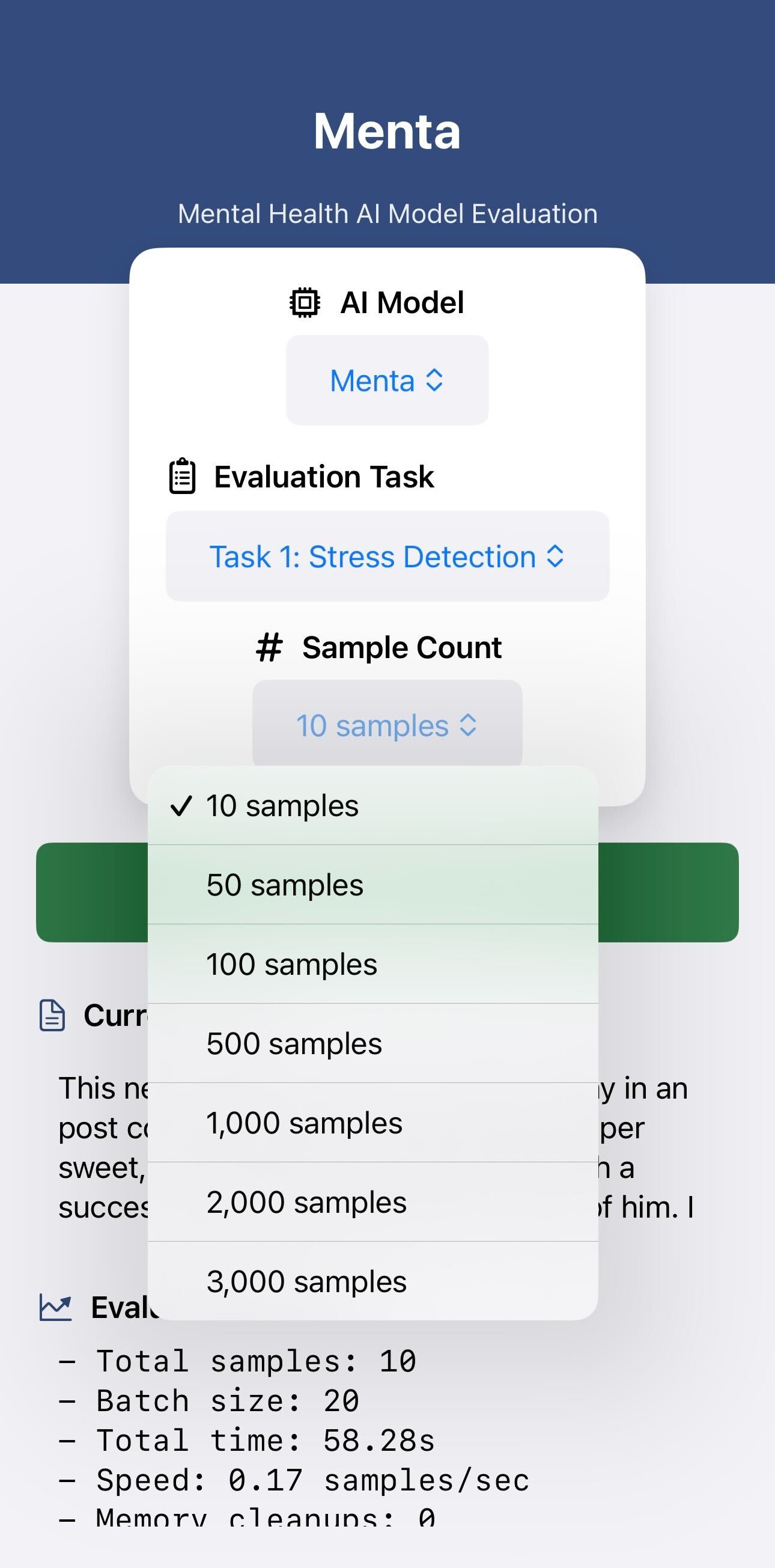}
  \caption{On-device deployment demonstration of Phi-4 Mini, Qwen-3, and Menta across six mental health tasks. The interface shows input social media posts and model predictions for model selection (left), task selection (middle), and sample count selection (right).}
  \label{fig:on-device-demos}
\end{figure}

\subsection{Deployment} \label{sec:deployment}

The development of the on-device mental health evaluation system demonstrates the feasibility of running multi-task mental health prediction entirely on-device, without requiring internet connectivity or exposing sensitive user data. Comparing deployment results among the three language models, Phi-4 Mini, Qwen-3, and Menta (Table \ref{tab:deploy_result}), we observe that Phi-4 Mini consistently shows the lowest TTFT across all tasks, suggesting better responsiveness (low latency) at the start of generation, using the least RAM (2.58–2.90 GB). Meanwhile, Qwen-3 generally achieves the highest ITPS, indicating faster token generation once decoding starts, though its OTPS is lower due to longer overall execution times in Task 5 and Task 6. The fine-tuned Menta offers a balanced performance, slightly slower in TTFT compared to Phi-4 Mini, but often competitive or superior in OTPS, especially in longer tasks (Tasks 5 and 6). Notably, Menta achieves up to 4$\times$ higher ITPS compared to existing SLMs such as Phi-4 Mini, while maintaining competitive output quality. It also achieves up to 4$\times$ faster total execution time compared to models like StableLM, with comparable memory usage. Additionally, Menta shows moderate RAM consumption ($\sim$3.0 GB), slightly higher than Phi-4 Mini but well within practical deployment limits.

\begin{figure}[t]
  \centering
  {
    \includegraphics[width=0.9\textwidth]{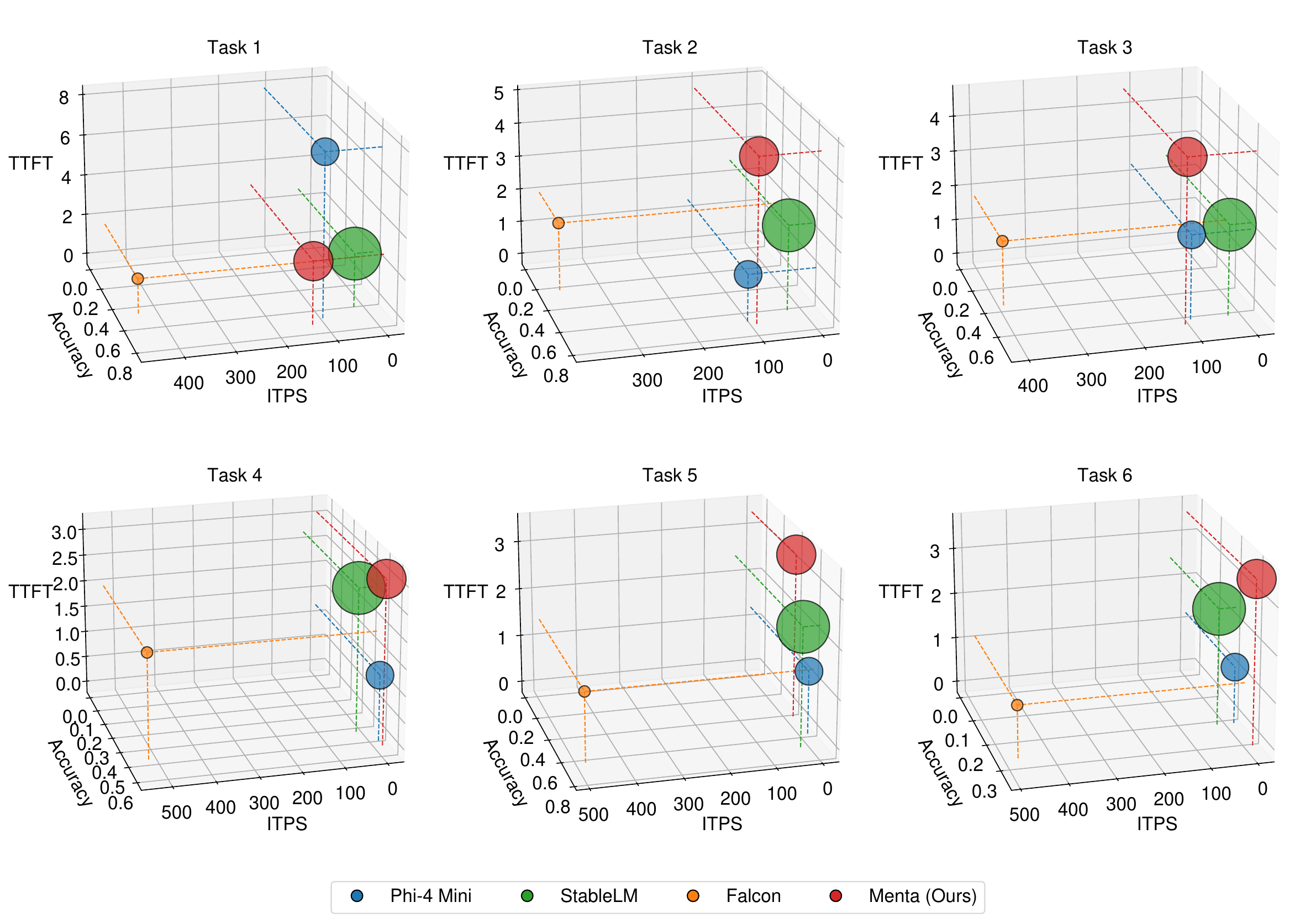}
  }
  \caption{A four-dimensional performance presentation for models Phi-4 Mini, StableLM, Falcon and Menta across ITPS in x axis, Accuracy in y axis, TTFT in z axis, and RAM (bubble size) in the deployment setting on an iPhone 15pro max device.}
  \label{fig:deployment_scatter}
\end{figure}

As shown in Figure \ref{fig:on-device-demos}, the Menta interface supports multiple on-device classification tasks, including stress, depression (both binary and severity-level), and suicide risk (both binary and categorical). Upon clicking the Start Evaluation button, social media posts will be input directly into the selected model, and the system will display the predicted class labels, correctness, and associated evaluation results. Model performance was assessed across all six tasks, with dataset sizes ranging from 10 to 3,000 posts depending on task complexity and availability.

Furthermore, our analysis reveals a clear trade-off between model efficiency and predictive performance when comparing Menta with other SLMs. From a deployment perspective, Menta offers a highly balanced accuracy, inference speed, and latency, positioning it as an ideal model for real-world applications where both performance and responsiveness are critical.

As illustrated in Figure~\ref{fig:deployment_scatter}, the three axes represent ITPS, TTFT, and accuracy. Each circle corresponds to a model, and the bubble size reflects memory consumption (RAM). Falcon prioritizes speed with fast TTFT and high ITPS, but at the expense of accuracy, achieving the lowest predictive performance among all models. Conversely, StableLM attains higher accuracy but incurs substantial latency and resource usage, resulting in increased deployment costs and slower user response times.

In contrast, Menta maintains top-tier accuracy (with the exception of Task 5, where it is narrowly outperformed by StableLM and Phi-4 Mini) while simultaneously achieving relatively lower latency of TTFT and competitive throughput of ITPS. This indicates that Menta generates high-quality outputs efficiently and promptly, which is essential for scalable, user-facing systems. Overall, while all models exhibit trade-offs across evaluation metrics, Menta strategically occupies an optimal deployment-ready position, delivering strong performance with manageable computational requirements, thereby reducing infrastructure overhead without compromising user experience.

\subsection{Case Study} \label{sec:case-study}

In this section, case studies are presented to underscore the key reasons in SLMs for nuanced mental health detection. 

\begin{figure}[t]
  \centering
  {
    \includegraphics[width=\textwidth, trim=35 0 15 0, clip]{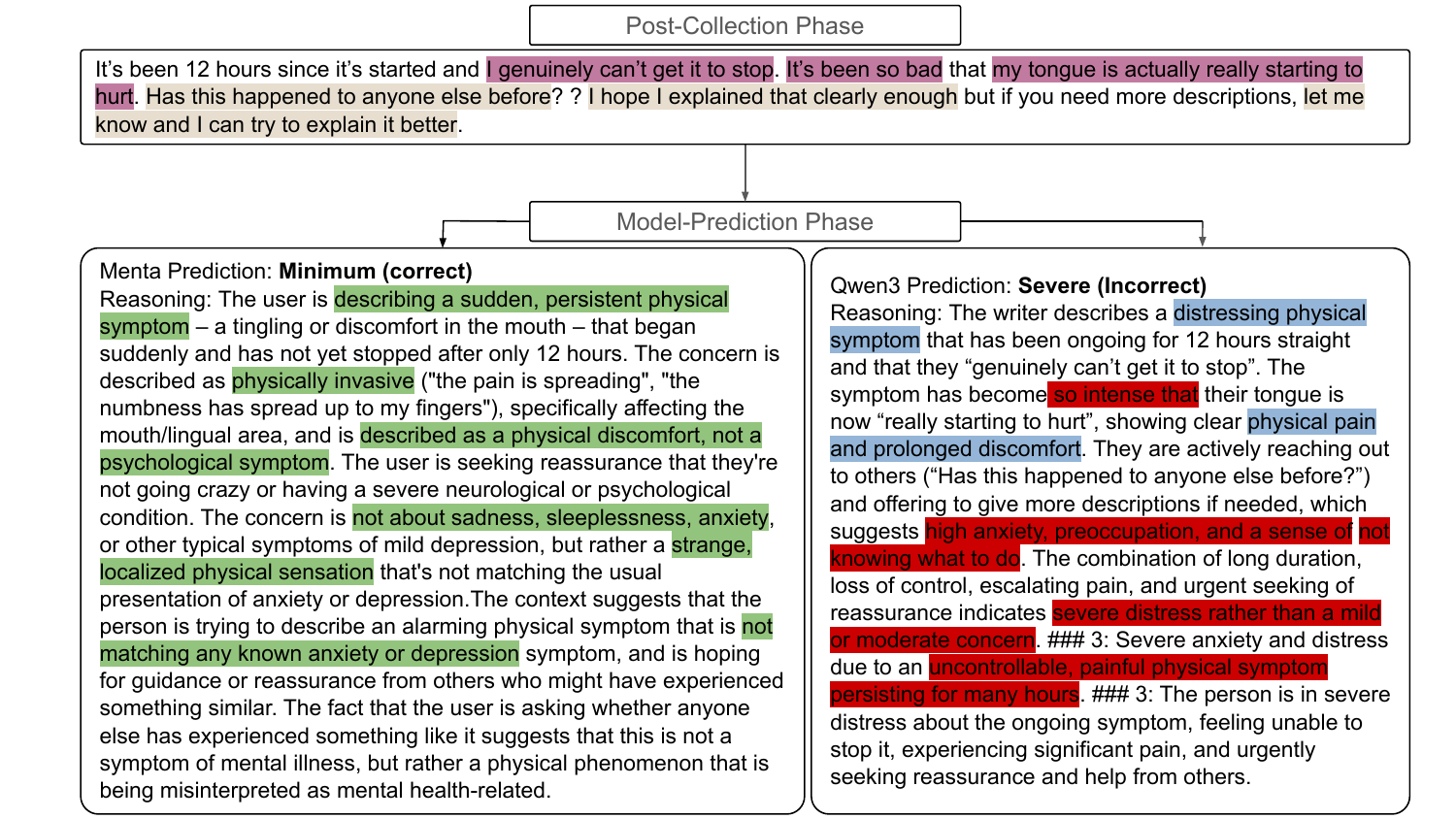}
  }
  \caption{Case 1: A successful prediction by the Menta model on Task 3 (depression ordinal classification), in contrast of Qwen-3’s misclassification of the same post. In the original post, expression of personal distress is highlighted in \colorbox{custompink}{pink}, while help-seeking and uncertainty is highlighted in \colorbox{custompink}{yellow}. In the SLM reasoning, correct analysis is marked in \colorbox{customgreen}{green}, correct analytical direction in \colorbox{customblue}{blue}, and overanalysis in \colorbox{customred}{red}.
}
  \label{fig:case_study1}
\end{figure}

\textbf{Case 1: Depression Level Detection}. In this instance, the Reddit post labeled as `Minimum', describes a user’s reflections on an involuntary and persistent action or sensation involving the tongue, as shown in Figure \ref{fig:case_study1}. The user expresses their concern and seek help toward a community, and they are willing to provide more details indicating a cooperative and communicative stance despite the discomfort. 

The Menta model correctly classifies this post as `Minimum' and offers a rationale consistent with that label. It identifies the user’s measured tone and composed approach to describing a physical issue, recognizing the absence of impulsive or emotionally escalated language as indicative of low psychological distress. This reflects a nuanced understanding that not all discomfort signals depression.

In contrast, Qwen-3 misclassifies the post as `Severe', interpreting the intensity of physical symptoms and expressions of discomfort as proxies for psychological severity. Its rationale conflates acute physical pain with anxiety and psychological distress, ultimately leading to an incorrect prediction. This failure stems from an overreliance on surface-level distress cues without differentiating somatic complaints from depressive features.

With fine-tuning, the Menta model has learned to focus on depression-specific markers and to align predictions with the annotation conventions of the dataset. This case study illustrates Menta’s ability to distinguish between contextual distress and clinical depression, highlighting its strength in predicting mental health states with greater precision.

\begin{figure}[t]
  \centering
  {
    \includegraphics[width=\textwidth, trim=35 25 15 50, clip]{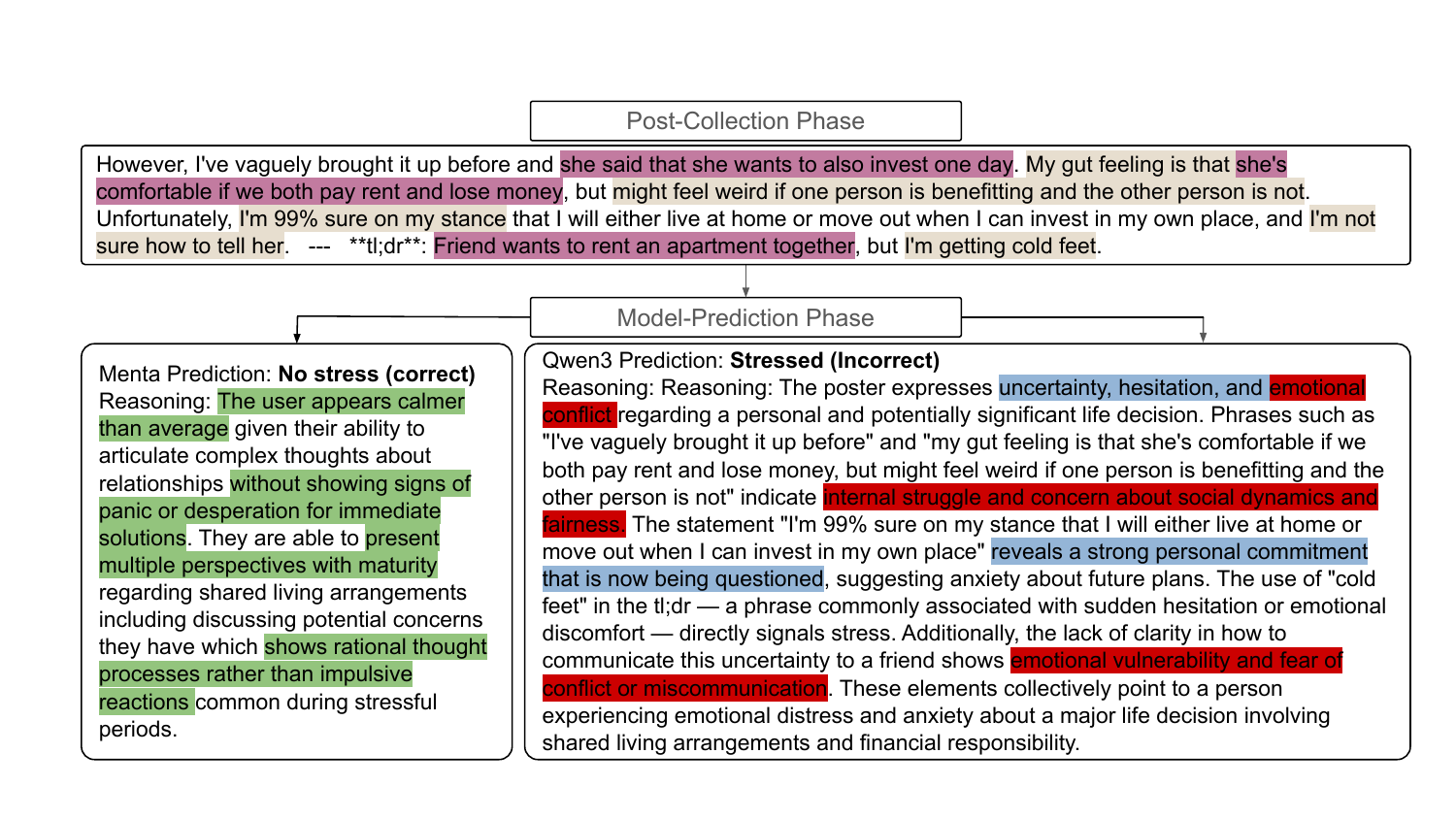}
  }
  \caption{Case 2: A successful prediction by the Menta model on Task 1 (stress classification), in contrast of Qwen-3’s misclassification of the same post. In the original post, cognitive or emotional conflict and decision-making stress are highlighted in \colorbox{custompink}{pink}, while interpersonal or social strain involving external relational dynamics is highlighted in \colorbox{customyellow}{yellow}. In the SLM’s reasoning, correct analysis is marked in \colorbox{customgreen}{green}, correct analytical direction in \colorbox{customblue}{blue}, and overanalysis in \colorbox{customred}{red}.
}
  \label{fig:case_study2}
\end{figure}

\textbf{Case 2: Stress Level Detection}. A Reddit post labeled as `No stress' describes a user’s reflections on a housing decision involving a friend. The user expresses thoughts calmly, lays out options rationally, and shows no indication of panic or overwhelming pressure.

The Menta model successfully classifies this post as `No stress' and provides a rationale aligned with this judgment. It highlights the user’s measured tone and maturity in navigating a potentially sensitive decision, correctly interpreting the absence of impulsive or anxious language as indicative of low stress. In contrast, Qwen-3 incorrectly predicts `Stressed', over-indexing on phrases such as `cold feet' and `99\% sure', and misinterpreting thoughtful deliberation as internal turmoil. According to the reasoning, Qwen-3 highlights the user’s stress, emotional discomfort, and indecision, which are psychological signals a well-functioning model should consider. However, it emphasizes external relational dynamic, social tension or potential relational mismatch, often linked to conflict-avoidant behavior or suppressed distress (highlighted red in Figure \ref{fig:case_study2}) and concludes with an incorrect prediction. Overall, this case study underscores the Menta model's ability to effectively predict mental health status with training.

\begin{figure}[t]
  \centering
  {
    \includegraphics[width=\textwidth, trim=35 60 15 55, clip]{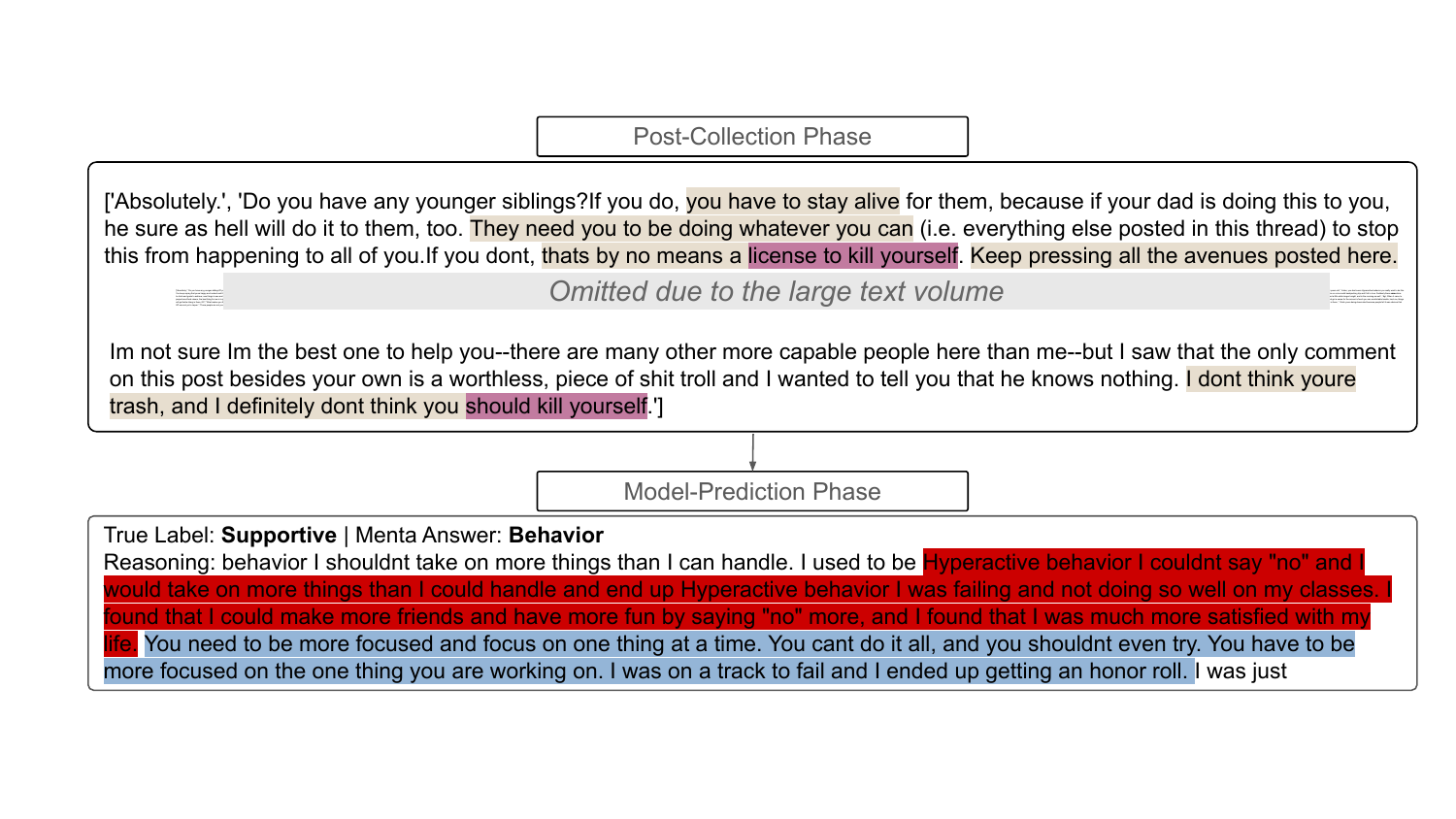}
  }
  \caption{Case 3: A failure case of the Menta model on Task 6 (suicide risk classification), where supportive or encouraging language is highlighted in \colorbox{customyellow}{yellow} and misinterpreted phrases in \colorbox{custompink}{pink} in the original post. Repeated ``behavior'' noise is highlighted in \colorbox{customred}{red}, while truly supportive content (advice, encouragement, and personal experience) highlighted in \colorbox{customblue}{blue}.}
  \label{fig:case_study3}
\end{figure}

\textbf{Case 3: Suicidal Risk Level Detection}. This case highlights a misclassification where a Reddit post with supportive level of suicidal risk was predicted as behavior by the Menta model (Figure \ref{fig:case_study3}). The primary cause was the excessive input length, which included a long multi-speaker thread with formatting noise (e.g., `\&gt';, list syntax, and HTML artifacts) and placeholder tokens such as `Hyperactive behavior'. These tokens acted as spurious cues, biasing the model toward behavior-related labels while ignoring clear supportive content such as `Please don’t kill yourself' or `Go to the police'.

The model’s training distribution includes both single-speaker inputs and multi-user interactions. In this case, the post aggregates interleaved content from multiple speakers without clear boundaries, increasing contextual complexity and deviating from typical training patterns. This structural mismatch, along with the extended input length, can cause the model to fixate on early salient cues while overlooking the broader supportive intent due to long-context instability.

\section{Discussion}

With task-specific supervision and systematic fine-tuning, our model Menta achieves strong performance across both binary and multi-level mental health classification tasks, including stress detection, depression severity, and suicidal ideation and risk assessment, capable of handling a diverse range of mental health outcomes within a unified, efficient framework. Our Menta model performs competitively with significantly larger LLMs, especially in depression and stress tasks, where it even surpasses larger models in certain settings. These results demonstrate the practical viability of well-optimized SLMs for continuous mental health monitoring, offering a cost-effective yet high-performing solution for early detection in digital mental health applications. In this section, we address three key questions regarding the use of SLMs and our fine-tuned Menta model in predicting mental health outcomes, while also outlining current limitations and directions for future research.

\textbf{\emph{Can SLMs effectively predict mental health status?}} Our findings demonstrate that SLMs possess substantial capability in predicting mental health conditions from text-based social media data. Despite their smaller parameter sizes, the well-trained Menta successfully handled complex classification tasks with great performance, which highlights the potential of such models to deliver accurate mental health predictions, supporting their use in low-resource or deployment-constrained settings.

While a previous study \citep{jia2025beyond} exploring SLMs in mental health focuses on SLMs in the 2–7B range, our experiments specifically target models strictly smaller than 7B and show that models below about 3.8B parameter scale (e.g., around 1B) fail to meaningfully perform, typically outputting blank or random results in zero- or few-shot settings. This suggests that scaling language models below a certain parameter threshold results in a catastrophic drop in task performance rather than a gradual degradation, indicating the presence of a sharper lower-bound capacity limit. Our work identifies this threshold explicitly within the mental health prediction domain. This observation is consistent with prior findings \citep{sengupta2025compression,subramanian2025small}, that performance degradation is not always smooth when size is reduced.

Our findings also align with prior work of MentalQLM (0.5B) \citep{shi2025mentalqlm}, which demonstrated strong performance in predicting a range of mental health outcomes including depression, stress, and suicidal ideation. These results highlight the viability of fine-tuned SLMs for mental health prediction, offering a favorable trade-off between computational efficiency and predictive accuracy. Moreover, consistent with recent studies \citep{ren2024learn, yang2024unveiling}, we observe that core reasoning and question-answering abilities are preserved after fine-tuning, enabling the resulting models to serve as multi-functional tools with superior performance on mental health tasks.

In summary, our experiments show that SLMs in the 3–7B parameter range offer a strong balance between efficiency and predictive performance with fine-tuning. While smaller models fail to generalize or recognize task constraints, models within this range effectively classify psychological signals in both zero-shot and few-shot settings. With targeted fine-tuning, our model Menta establishes a clear lower bound for model size, achieving strong performance on complex mental health tasks while maintaining computational efficiency.

\textbf{\emph{Balancing model size, accuracy, safety, and real-world impact in mental health AI}}. Comparing the performance of SLMs and larger LLMs, fine-tuned SLMs show comparable or superior performance in depression and stress level classification tasks while lower performance in suicidal ideation prediction tasks. However, compared with non-trained SLMs, the performance improves in a large degree as well as retaining a more balanced accuracy among prediction tasks and classes. Our findings show that that well-optimized lightweight SLMs can retain core reasoning and contextual understanding necessary for mental health inference. 

Our findings were observed in previous work \citep{jia2025beyond} that SLMs often came within 2\% of LLMs’ F1 scores in binary classification settings under zero-shot evaluation, highlighting the potential of SLMs as resource-efficient and privacy-conscious alternatives in clinical or low-resource environments. Other prior works have concentrated primarily on LLMs or hybrid models in mental health domains. For instance, Kim et al. \citep{kim2025interpretable} compares zero-shot LLMs and supervised classifiers built on LLM embeddings, showing that while LLMs generalize well in binary depression classification, they struggle in fine-grained severity classification tasks, where our findings collaborate the difficulty in finer labels. Recent surveys \citep{ge2025survey,garg2025rise} highlight the growing promise of SLMs in digital health applications, noting their efficiency, reduced computational demands, and capacity for local deployment, which offer significant advantages over LLMs by being more privacy-preserving and easier to integrate into real-world clinical or mobile settings. Our work complements these surveys by providing empirical evidence for the operational boundary of this trade-off, showing precisely when SLMs transition from viable to non-viable performance levels for such sensitive mental health applications. Also, our compact Menta model demonstrates that the on-device SLM deployment is essential for protecting user privacy, a primary ethical concern in digital psychiatry \citep{patel2018lancet,jin2025applications}. In resource-limited clinical settings, particularly in low- and middle-income countries where mental-health workforce shortages persist \citep{moitra2022global}, SLMs enable accessible, cost-efficient, and scalable mental-health screening tools. Moreover, smaller models are easier to interpret and audit, aligning with recent calls for transparent and accountable AI in healthcare \citep{doshi2017towards,tonekaboni2019clinicians}. 

The competitiveness of SLMs in this domain primarily emerges because mental-health language tasks often rely on localized lexical, affective, and syntactic cues, such as expressions of self-reference, emotional polarity, and temporal framing, which can be effectively captured by moderate-sized models without requiring the broad world knowledge encoded in LLMs. As shown in prior work \citep{ji2021mentalbert,shen2018cross,benton2017multi}, fine-tuning on domain-specific corpora substantially narrows the performance gap between smaller and larger architectures. Our findings extend this evidence by demonstrating that parameter-efficient fine-tuning (e.g., LoRA) combined with adaptive, weighted loss functions enables SLMs to learn discriminative representations for mental-health prediction at a fraction of the computational cost.

While LLMs generally outperform across a broader range of tasks due to their larger capacity and instruction-following capabilities, SLMs demonstrate competitive and even superior performance on specific tasks, particularly depression and stress prediction tasks. For the remaining tasks, although SLMs show notable improvements, their performance still underperforms compared with LLMs \citep{xu2024mental}. This suggests that SLMs can be highly effective when the task structure and linguistic signals are well-aligned with their capacity. However, a critical limitation remains: SLMs struggle with long input sequences due to both token-length constraints and reduced capability in handling extended contextual dependencies \citep{kumar2025large}. This restricts their applicability in scenarios requiring nuanced understanding of lengthy or complex user narratives, where LLMs maintain a distinct advantage \citep{lu2024small}.

Overall, our findings support a paradigm shift from size-centric scaling to purpose-centric optimization in mental-health AI. As efficiency, transparency, safety and user privacy \citep{zhang2021breaking} become increasingly prioritized in clinical machine learning, the balance between performance and practicality will define the next generation of digital mental-health tools. SLMs address emerging demands in digital mental health for in-situ, low-latency, and privacy-preserving early detection \citep{harari2016using, wang2018tracking}, while also aligning with growing calls for transparent and accountable AI in healthcare settings \citep{wies2021digital, smith2023digital}. By showing that fine-tuned SLMs with adaptive, cross-dataset strategies can match or surpass LLMs in mental-health outcome prediction, this study provides empirical evidence that smaller, well-aligned models may offer a more sustainable and ethically sound path forward for real-world mental-health monitoring.

\textbf{\emph{What tasks are most suitable for SLMs?}} In our work, we demonstrate that cross-dataset training is not only feasible but beneficial for predicting depression, stress, and suicidal ideation, when combined with an adaptive loss design. This finding resonates with observations from Yao et al. \citep{benton2017multi}, who reported that shared latent representations across related affective tasks can promote positive transfer. However, unlike their multi-task architecture, our approach achieves comparable cross-task gains within a parameter-efficient fine-tuning paradigm (LoRA), thereby reducing computational cost and memory footprint. This distinction highlights that even lightweight fine-tuning frameworks can leverage dataset diversity effectively, provided the optimization objective is properly calibrated.

Also, we found that the fine-tuned model Menta shows better performance in depression and stress detection tasks compared with suicidal risk detection. We found that fine-tuned and non-fine-tuned SLMs, and even larger language models, tend to achieve stronger performance on depression and stress detection tasks compared to suicide risk classification. For example, a systematic review \citep{bauer2024using} further notes that language models often misclassify nuanced expressions of suicidality due to class imbalance and the reliance on contextual or temporal dependencies, which are not captured in single posts. Our findings also collaborate with another study \citep{lamichhane2023evaluation}, which reported F1‑scores of 0.86 for depression detection and only 0.37 for suicidality detection on social media data. These findings reflect that depression and stress often manifest with clearer linguistic patterns and more abundant data, whereas suicidality may be rarer, more context dependent, and linguistically subtler, making it harder for models to learn reliably. As a consequence, SLMs (and even larger models) may be better suited for screening tasks (e.g., detecting depression or stress) than for nuanced stratification of suicide risk without additional multimodal or temporal data.

\textbf{\emph{Limitation and future directions}}. Despite demonstrating that SLMs such as Qwen-3 (4B) can effectively perform mental health prediction tasks, several limitations constrain the generalizability and interpretability of our findings.

While our evaluation used diverse datasets covering depression, stress, and suicidal ideation, all inputs were textual social media posts. This focus excludes multimodal features such as linguistic style dynamics, user metadata, or interaction patterns, that often play a critical role in understanding real-world mental health contexts. Incorporating multimodal signals could improve robustness and ecological validity in future work \citep{yazdavar2020multimodal,khoo2024machine}. Also, one key limitation of this study lies in the restricted diversity of datasets and mental health tasks explored. Our evaluation primarily focused on three task categories depression, stress, and suicidal ideation, each derived from publicly available social media datasets. While these benchmarks are widely adopted in affective computing, they still represent a narrow subset of the psychological spectrum. Real-world mental health expressions often encompass more nuanced and overlapping conditions such as anxiety, bipolar disorder, and emotional dysregulation, which are underrepresented in our current corpus \citep{cao2024machine}. Additionally, all analyses were conducted on publicly available, anonymized datasets rather than in clinical environments. While this design ensures privacy and reproducibility, it limits conclusions about clinical safety, ethical compliance, and longitudinal reliability \citep{seyedsalehi2024suicide}. Future directions include the test of federated or on-device fine-tuned SLMs in clinical settings. Deploying such models as assistive rather than diagnostic tools, under clinician supervision, could provide a safe bridge between algorithmic insights and responsible mental health support.

\section{Conclusion}

In conclusion, this work demonstrates the feasibility and effectiveness of SLMs for digital mental health prediction from social media text. By introducing Menta, the first optimized and compact SLM fine-tuned across six mental health tasks, we show that careful architectural and training design enables high performance without sacrificing efficiency or deployability. Menta not only achieves strong predictive accuracy, especially for binary stress and suicidality detection, but also operates in real time on resource-constrained devices such as smartphones. These findings highlight that SLMs can offer scalable, private, and interpretable mental health support, bridging the gap between clinical relevance and technological accessibility.

These findings addresses critical challenges of cost, accessibility, and privacy that limit traditional approaches. While our results provide encouraging evidence, further work should evaluate multimodal and cross-lingual extensions, clinical validation, and interpretability to ensure responsible translation into practice. Overall, Menta demonstrates that small, task-optimized language models can deliver accurate, efficient, and interpretable mental health predictions, while enabling scalable and privacy-preserving real-world deployment and shows that adaptive fine-tuning can bridge the gap between efficiency and performance in next-generation digital mental-health systems.

\bibliography{iclr2026_conference}
\bibliographystyle{iclr2026_conference}

\appendix
\section{Appendix}

\begin{figure}[htbp]
  \centering
  {
    \includegraphics[width=1.0\textwidth, trim=50 100 45 15, clip]{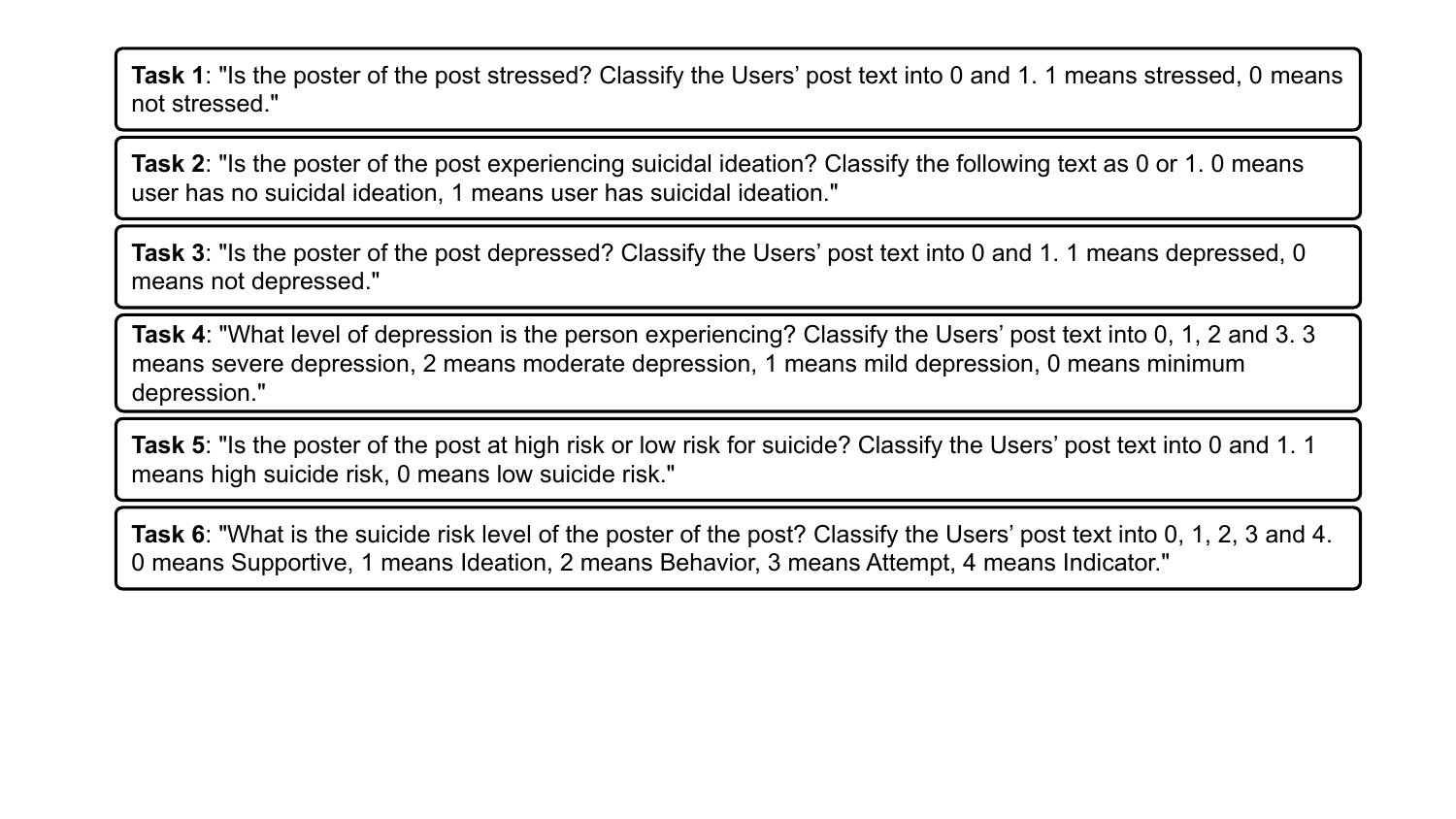}
  }
  \caption{Prompt Instruction for the Six Mental Health Tasks.}
  \label{fig:task_prompts}
\end{figure}

\begin{figure}[htbp]
  \centering
  \subfloat[Menta-T1]{%
    \includegraphics[width=0.32\textwidth]{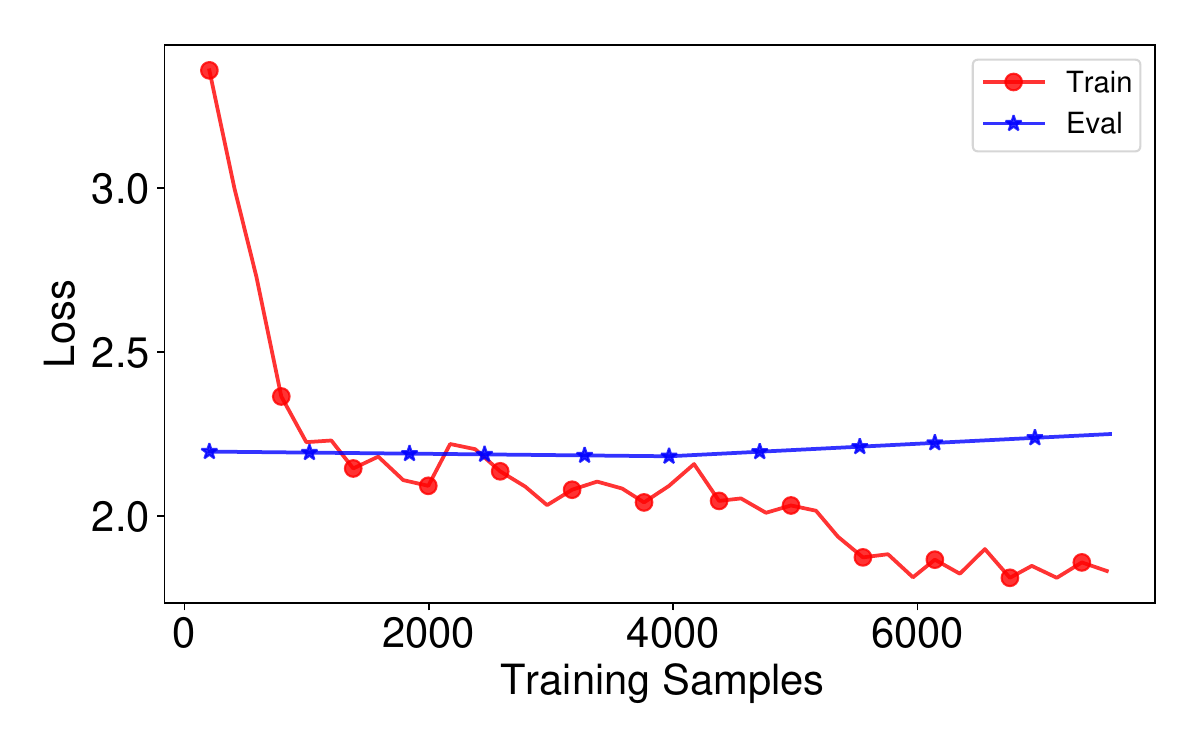}%
    \label{fig:task1_loss}%
  }\hfill
  \subfloat[Menta-T2]{%
    \includegraphics[width=0.32\textwidth]{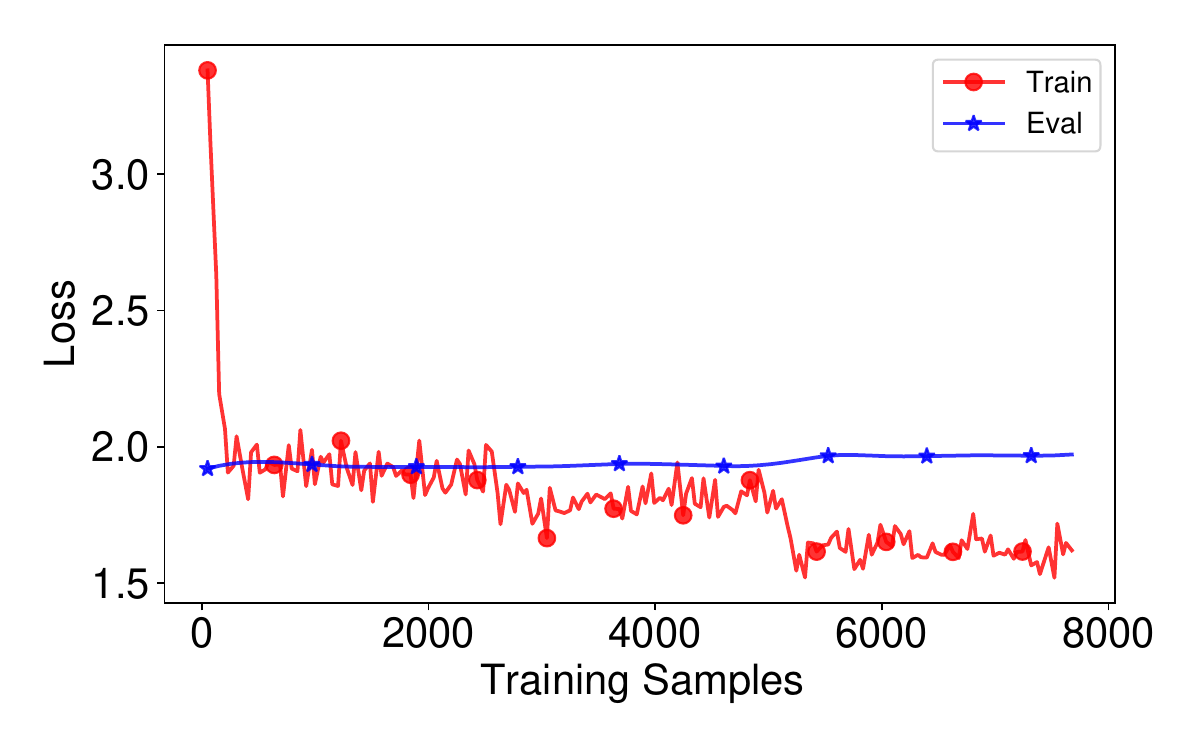}%
    \label{fig:task2_loss}%
  }\hfill
  \subfloat[Menta-T3]{%
    \includegraphics[width=0.32\textwidth]{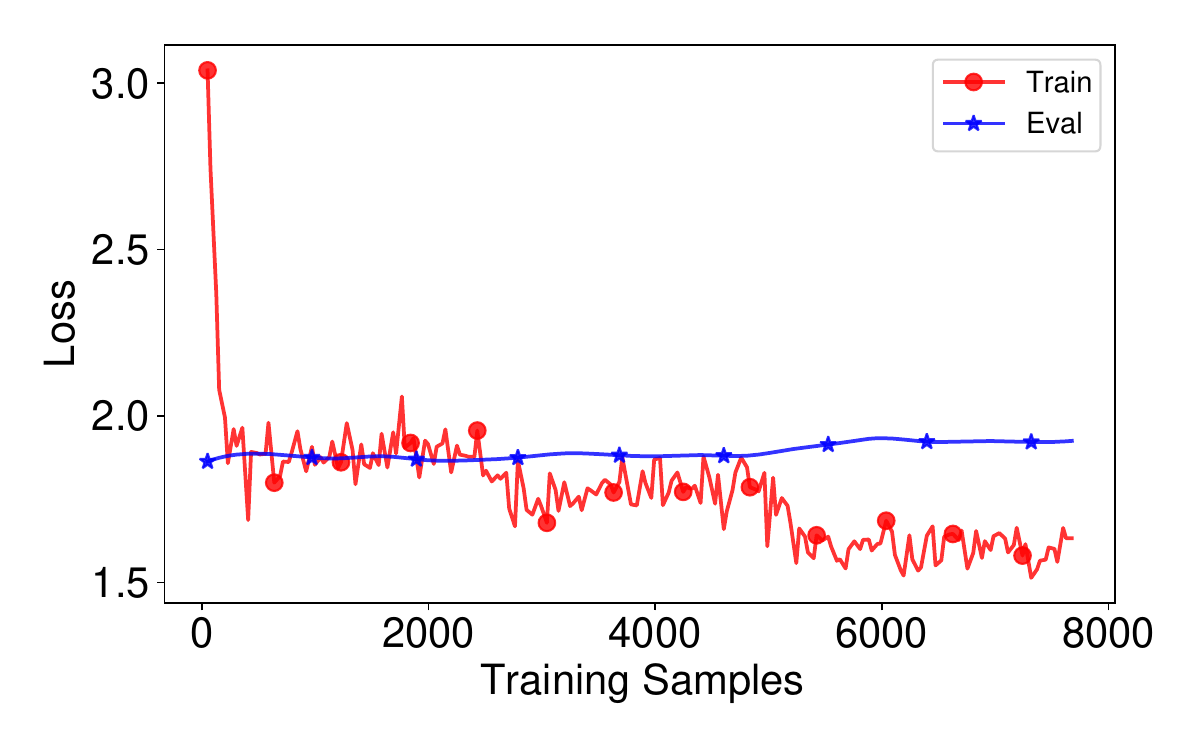}%
    \label{fig:task3_loss}%
  }\\[4pt] 

  \subfloat[Menta-T4]{%
    \includegraphics[width=0.32\textwidth]{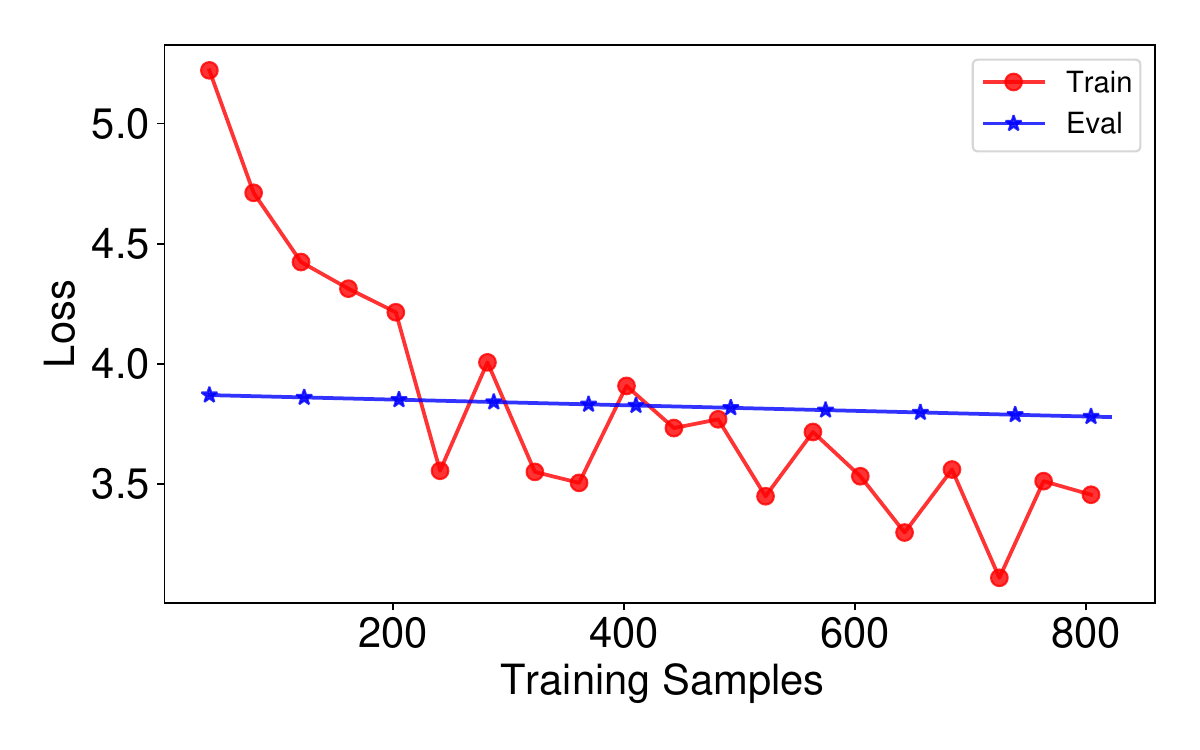}%
    \label{fig:task4_loss}%
  }\hfill
  \subfloat[Menta-T5]{%
    \includegraphics[width=0.32\textwidth]{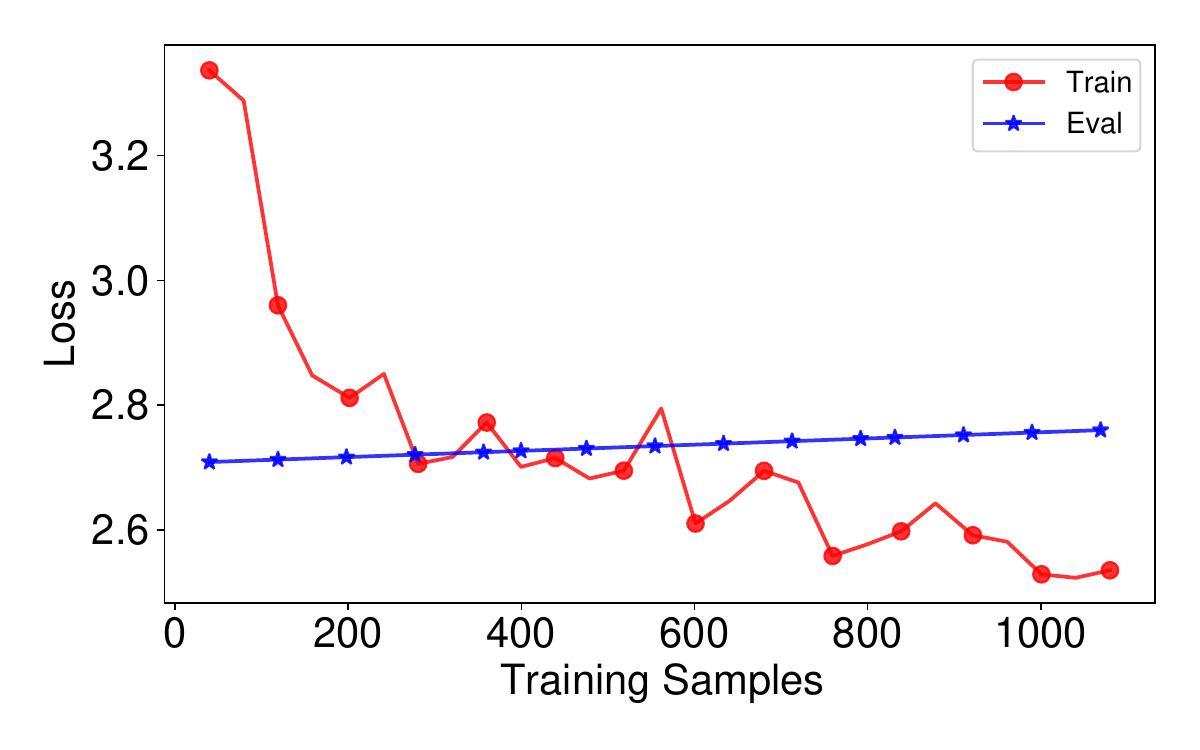}%
    \label{fig:task5_loss}%
  }\hfill
  \subfloat[Menta-T6]{%
    \includegraphics[width=0.32\textwidth]{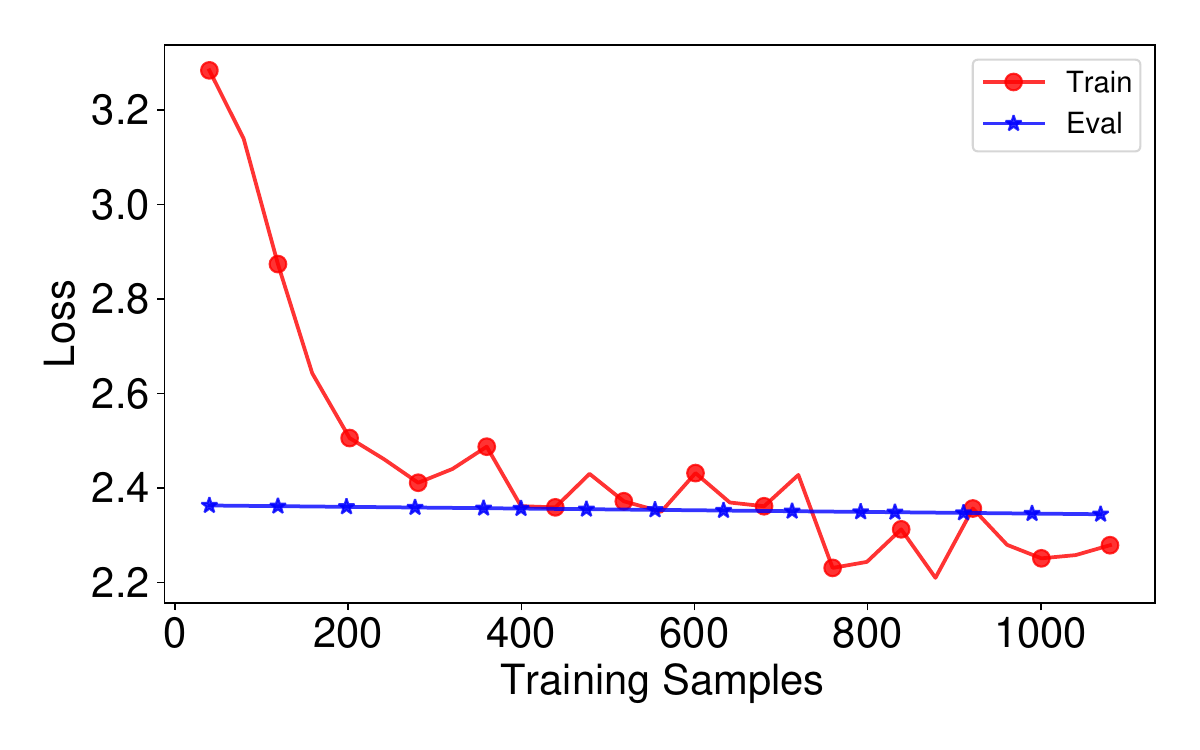}%
    \label{fig:task6_loss}%
  }\\[4pt]

  \subfloat[Menta (general)]{%
    \includegraphics[width=0.37\textwidth]{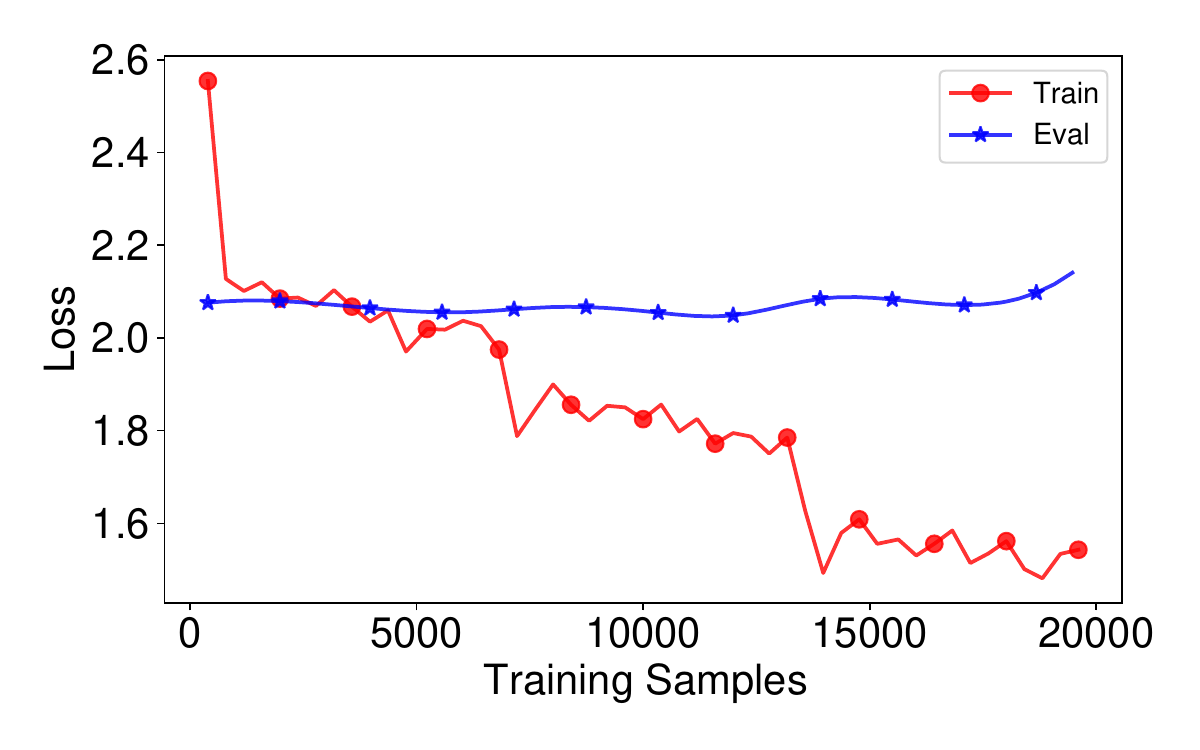}%
    \label{fig:menta_loss}%
  }

  \caption{Training loss for the six individually fine-tuned models trained with Task 1-6 data respectively, compared with the Menta model trained on all six tasks.}
  \label{fig:loss}
\end{figure}

\begin{table*}[htbp]
\centering
\scriptsize
\caption{Accuracy of different models across tasks and few-shot settings (mean ± standard deviation), with the best performance highlighted in bold and the second-best underlined.}
\resizebox{\textwidth}{!}{
\begin{tabular}{llccccc}
\toprule
Task & Model-Size & Zero-shot & One-shot & Two-shot & Three-shot & Four-shot \\
\midrule

\multirow{7}{*}{Task 1} 
& Gemma-3 (1B) & 0.48 $\pm$ 0.01 & 0.53 $\pm$ 0.03 & 0.55 $\pm$ 0.02 & 0.56 $\pm$ 0.03 & 0.52 $\pm$ 0.02 \\
& Gemma-3 (4B) & 0.53 $\pm$ 0.03 & 0.60 $\pm$ 0.05 & 0.57 $\pm$ 0.03 & 0.59 $\pm$ 0.02 & 0.58 $\pm$ 0.03 \\
& Qwen-3 (4B) & \underline{0.56} $\pm$ 0.01 & \underline{0.64} $\pm$ 0.04 & \textbf{0.62} $\pm$ 0.02 & \underline{0.63} $\pm$ 0.02 & \textbf{0.62} $\pm$ 0.02 \\
& Phi-4 Mini (3.8B) & \textbf{0.57} $\pm$ 0.02 & \textbf{0.69} $\pm$ 0.02 & 0.59 $\pm$ 0.02 & \textbf{0.66} $\pm$ 0.03 & \textbf{0.62} $\pm$ 0.02 \\
& TinyLLaMA (1.1B) & 0.50 $\pm$ 0.01 & 0.51 $\pm$ 0.01 & 0.51 $\pm$ 0.01 & 0.50 $\pm$ 0.05 & 0.52 $\pm$ 0.01 \\
& Falcon (1.3B) & 0.51 $\pm$ 0.01 & 0.51 $\pm$ 0.02 & 0.51 $\pm$ 0.01 & 0.51 $\pm$ 0.01 & 0.53 $\pm$ 0.01 \\
& StableLM (3B) & 0.50 $\pm$ 0.00 & 0.59 $\pm$ 0.00 & \underline{0.59} $\pm$ 0.00 & 0.59 $\pm$ 0.00 & \underline{0.59} $\pm$ 0.00 \\
\midrule

\multirow{7}{*}{Task 2} 
& Gemma-3 (1B) & 0.50 $\pm$ 0.00 & 0.57 $\pm$ 0.06 & 0.61 $\pm$ 0.08 & 0.53 $\pm$ 0.08 & 0.55 $\pm$ 0.03 \\
& Gemma-3 (4B) & 0.63 $\pm$ 0.01 & \underline{0.60} $\pm$ 0.08 & 0.63 $\pm$ 0.03 & 0.59 $\pm$ 0.01 & 0.62 $\pm$ 0.05 \\
& Qwen-3 (4B) & \textbf{0.70} $\pm$ 0.01 & \textbf{0.66} $\pm$ 0.03 & \textbf{0.67} $\pm$ 0.01 & \textbf{0.70} $\pm$ 0.02 & \textbf{0.67} $\pm$ 0.01 \\
& Phi-4 Mini (3.8B) & \underline{0.69} $\pm$ 0.01 & \textbf{0.66} $\pm$ 0.05 & \underline{0.66} $\pm$ 0.04 & \underline{0.68} $\pm$ 0.02 & \underline{0.66} $\pm$ 0.01 \\
& TinyLLaMA (1.1B) & 0.50 $\pm$ 0.00 & 0.50 $\pm$ 0.00 & 0.50 $\pm$ 0.01 & 0.50 $\pm$ 0.00 & 0.50 $\pm$ 0.01 \\
& Falcon (1.3B) & 0.50 $\pm$ 0.00 & 0.50 $\pm$ 0.00 & 0.50 $\pm$ 0.00 & 0.50 $\pm$ 0.02 & 0.50 $\pm$ 0.01 \\
& StableLM (3B) & 0.53 $\pm$ 0.01 & 0.51 $\pm$ 0.01 & 0.51 $\pm$ 0.01 & 0.51 $\pm$ 0.01 & 0.51 $\pm$ 0.01 \\
\midrule

\multirow{7}{*}{Task 3}
 & Gemma3 (1B) & 0.25 ± 0.01 & 0.29 ± 0.03 & 0.30 ± 0.03 & 0.28 ± 0.01 & 0.28 ± 0.01 \\
 & Gemma3 (4B) & \underline{0.30} ± 0.01 & \underline{0.32} ± 0.02 & 0.31 ± 0.01 & 0.30 ± 0.01 & 0.31 ± 0.02 \\
 & Qwen-3 (4B) & \underline{0.30} ± 0.00 & 0.33 ± 0.02 & \underline{0.34} ± 0.01 & \underline{0.34} ± 0.02 & \textbf{0.37} ± 0.01 \\
 & Phi-4 Mini (3.8B) & \textbf{0.35} ± 0.01 & \textbf{0.38} ± 0.01 & \textbf{0.38} ± 0.00 & \textbf{0.37} ± 0.01 & \textbf{0.37} ± 0.01 \\
 & TinyLLama (1.1B) & 0.25 ± 0.00 & 0.29 ± 0.00 & 0.29 ± 0.00 & 0.29 ± 0.00 & 0.29 ± 0.00 \\
 & Falcon (1.3B) & 0.24 ± 0.01 & 0.25 ± 0.00 & 0.26 ± 0.01 & 0.26 ± 0.01 & 0.26 ± 0.01 \\
 & StableLM (3B) & 0.24 ± 0.01 & 0.29 ± 0.00 & 0.29 ± 0.00 & 0.29 ± 0.00 & \underline{0.29} ± 0.00 \\
\midrule
\multirow{7}{*}{Task 4}
 & Gemma-3 (1B) & 0.36 ± 0.01 & 0.35 ± 0.02 & 0.35 ± 0.02 & 0.36 ± 0.01 & 0.34 ± 0.01 \\
 & Gemma-3 (4B) & 0.37 ± 0.01 & 0.37 ± 0.01 & 0.37 ± 0.01 & 0.36 ± 0.01 & 0.36 ± 0.01 \\
 & Qwen-3 (4B) & \underline{0.41} ± 0.01 & \underline{0.40} ± 0.01 & \underline{0.41} ± 0.01 & \underline{0.41} ± 0.01 & \underline{0.42} ± 0.01 \\
 & Phi-4 Mini (3.8B) & \textbf{0.42} ± 0.01 & \textbf{0.43} ± 0.01 & \textbf{0.43} ± 0.01 & \textbf{0.43} ± 0.01 & \textbf{0.43} ± 0.01 \\
 & TinyLLama (1.1B) & 0.29 ± 0.00 & 0.29 ± 0.00 & 0.30 ± 0.00 & 0.30 ± 0.00 & 0.30 ± 0.01 \\
 & Falcon (1.3B) & 0.27 ± 0.01 & 0.28 ± 0.00 & 0.29 ± 0.01 & 0.29 ± 0.00 & 0.28 ± 0.00 \\
 & StableLM (3B) & 0.27 ± 0.00 & 0.28 ± 0.00 & 0.28 ± 0.00 & 0.28 ± 0.00 & 0.28 ± 0.00 \\
\midrule
\multirow{7}{*}{Task 5}
 & Gemma-3 (1B) & 0.46 ± 0.01 & 0.49 ± 0.01 & 0.48 ± 0.03 & 0.49 ± 0.01 & 0.50 ± 0.03 \\
 & Gemma-3 (4B) & 0.48 ± 0.00 & \underline{0.54} ± 0.02 & \underline{0.51} ± 0.02 & \underline{0.53} ± 0.01 & \underline{0.54} ± 0.02 \\
 & Qwen-3 (4B) & \underline{0.53} ± 0.01 & 0.51 ± 0.01 & 0.49 ± 0.03 & 0.49 ± 0.01 & 0.49 ± 0.01 \\
 & Phi-4 Mini (3.8B) & \textbf{0.58} ± 0.01 & \textbf{0.58} ± 0.02 & \textbf{0.61} ± 0.02 & \textbf{0.60} ± 0.01 & \textbf{0.59} ± 0.01 \\
 & TinyLLama (1.1B) & 0.50 ± 0.00 & 0.46 ± 0.01 & \underline{0.51} ± 0.02 & 0.49 ± 0.02 & 0.48 ± 0.01 \\
 & Falcon (1.3B) & 0.50 ± 0.00 & 0.50 ± 0.00 & 0.50 ± 0.00 & 0.50 ± 0.00 & 0.50 ± 0.00 \\
 & StableLM (3B) & 0.46 ± 0.02 & 0.50 ± 0.01 & 0.49 ± 0.02 & 0.50 ± 0.01 & 0.51 ± 0.02 \\
\midrule
\multirow{7}{*}{Task 6}
 & Gemma-3 (1B) & 0.20 ± 0.01 & 0.20 ± 0.00 & 0.20 ± 0.01 & 0.20 ± 0.01 & 0.20 ± 0.01 \\
 & Gemma-3 (4B) & \underline{0.22} ± 0.01 & 0.24 ± 0.02 & 0.24 ± 0.01 & \textbf{0.26} ± 0.01 & \textbf{0.27} ± 0.01 \\
 & Qwen-3 (4B) & \textbf{0.23} ± 0.12 & \textbf{0.26} ± 0.02 & \textbf{0.27} ± 0.02 & \textbf{0.26} ± 0.02 & \textbf{0.27} ± 0.01 \\
 & Phi-4 Mini (3.8B) & 0.21 ± 0.00 & \textbf{0.26} ± 0.00 & \underline{0.26} ± 0.01 & \textbf{0.26} ± 0.00 & \underline{0.24} ± 0.03 \\
 & TinyLLama (1.1B) & \underline{0.22} ± 0.02 & 0.21 ± 0.01 & 0.21 ± 0.01 & 0.21 ± 0.01 & 0.21 ± 0.01 \\
 & Falcon (1.3B) & 0.20 ± 0.00 & 0.20 ± 0.01 & 0.19 ± 0.01 & 0.20 ± 0.01 & 0.21 ± 0.02 \\
 & StableLM (3B) & 0.20 ± 0.00 & \underline{0.25} ± 0.00 & 0.24 ± 0.00 & \underline{0.24} ± 0.01 & \underline{0.24} ± 0.00 \\
\bottomrule
\end{tabular}
}
\label{tab:fewshot_results}
\end{table*}

\begin{table}[htbp]
\centering
\small
\caption{Balanced Accuracy (BACC) results across six tasks with mean and standard deviation, with the best performance highlighted in bold and the second-best underlined.}
\begin{tabular}{lcccccc}
\toprule
Model & Task 1 & Task 2 & Task 3 & Task 4 & Task 5 & Task 6 \\
\midrule
Menta-T1 & \textbf{0.83} ± 0.02 & 0.67 ± 0.04 & 0.30 ± 0.10 & \textbf{0.68} ± 0.04 & 0.55 ± 0.01 & 0.25 ± 0.01 \\
Menta-T2 & 0.79 ± 0.04 & \textbf{0.75} ± 0.04 & 0.31 ± 0.01 & 0.65 ± 0.04 & \underline{0.56} ± 0.02 & 0.23 ± 0.03 \\
Menta-T3 & 0.77 ± 0.02 & 0.66 ± 0.05 & \underline{0.37} ± 0.06 & \underline{0.66} ± 0.03 & 0.52 ± 0.02 & \textbf{0.26} ± 0.02 \\
Menta-T4 & 0.76 ± 0.01 & 0.63 ± 0.04 & 0.27 ± 0.04 & 0.63 ± 0.03 & \underline{0.56} ± 0.01 & \underline{0.25} ± 0.02 \\
Menta-T5 & 0.58 ± 0.02 & 0.55 ± 0.04 & 0.29 ± 0.03 & 0.51 ± 0.00 & 0.47 ± 0.06 & 0.22 ± 0.03 \\
Menta-T6 & 0.66 ± 0.05 & 0.65 ± 0.05 & 0.27 ± 0.03 & 0.54 ± 0.02 & 0.47 ± 0.06 & 0.24 ± 0.04 \\
Menta (General Model) & \underline{0.81} ± 0.02 & \underline{0.73} ± 0.05 & \textbf{0.38} ± 0.07 & 0.61 ± 0.03 & \textbf{0.60} ± 0.03 & \textbf{0.26} ± 0.05 \\
\bottomrule
\end{tabular}
\label{tab:task_all_results}
\end{table}

\begin{table}[htbp]
\centering
\caption{Consolidated deployment performance across all tasks, with the best performance highlighted in bold and the second-best underlined.}
\label{tab:deploy_result}
\begin{tabular}{llcccccc}
\toprule
\textbf{Task} & \textbf{Model} & \textbf{TTFT (s)} & \textbf{ITPS (/s)} & \textbf{OET (s)} & \textbf{OTPS (/s)} & \textbf{Total Time (s)} & \textbf{RAM (GB)} \\
\midrule
\multirow{4}{*}{Task 1} & Phi-4 Mini        & \textbf{1.239} & 870.8  & 8.148  & \underline{5.4} & 5834.33  & \textbf{2.58} \\
                        & Qwen-3       & 3.705 & \underline{3516.5} & \textbf{5.193}  & 1.9 & \textbf{3718.19}  & \underline{2.94}  \\
                        & Menta & 4.024 & \textbf{3587.0} & \underline{6.335}  & 1.8 & \underline{4535.86}  & 2.99 \\
                        & StableLM	& \underline{2.599}	& 37.5	& 23.2	& \textbf{6.6}	& 16541.60	& 4.77 \\
\midrule
\multirow{4}{*}{Task 2} & Phi-4 Mini        & \textbf{1.376} & 723.1  & \underline{8.174}  & \underline{4.4} & \underline{29050.40} & \textbf{2.58} \\
                        & Qwen-3       & 4.202 & \textbf{2941.3} & \textbf{6.760}  & 1.6 & \textbf{24025.04} & \underline{3.03} \\
                        & Menta & 4.777 & \underline{2931.6} & 10.885 & 1.9 & 38685.29 & 3.04 \\
                        & StableLM	& \underline{2.486}	& 38.6	& 22.54	& \textbf{6.7}	& 80084.62	& 3.89 \\
\midrule
\multirow{4}{*}{Task 3} & Phi-4 Mini         & \textbf{2.333} & 1055.1 & \underline{8.384}  & \underline{5.1} & \underline{29668.79} & \textbf{2.58} \\
                        & Qwen-3       & 4.288 & \textbf{4337.2} & \textbf{6.834}  & 1.6 & \textbf{24265.29} & \underline{3.03} \\
                        & Menta & 5.083 & \underline{4275.4} & 12.837 & 1.8 & 45622.70 & 3.04 \\
                        & StableLM	& \underline{2.512}	& 38.3	& 22.288	& \textbf{6.8}	 & 79189.26	& 3.91 \\
\midrule
\multirow{4}{*}{Task 4} & Phi-4 Mini        & \textbf{1.236} & 1097.0 & 8.213  & \underline{3.4} & 3120.94  & \textbf{2.90} \\
                        & Qwen-3       & 3.042 & \textbf{2930.2} & \textbf{4.875}  & 1.5 & \textbf{1852.50}  & \underline{2.94} \\
                        & Menta & 3.080 & \underline{2698.1} & \underline{4.910}  & 1.4 & \underline{1917.20}  & 3.00 \\
                        & StableLM	& \underline{2.708}	& 33.5	& 24.473	& \textbf{6.3}	& 9299.74	& 4.11\\
\midrule
\multirow{4}{*}{Task 5} & Phi-4 Mini        & \textbf{1.287} & 1041.5 & \textbf{8.656}  & \underline{5.1} & \textbf{4328.00}  & \textbf{2.59} \\
                        & Qwen-3       & 6.966 & \underline{5550.0} & 25.428 & 1.3 & 12650.30 & \underline{2.95} \\
                        & Menta & 6.005 & \textbf{6031.4} & 27.657 & 1.3 & 13828.50 & 3.04 \\
                        & StableLM	& \underline{2.431}	& 37.4	& \underline{24.104}	& \textbf{6.3}	& \underline{12052.00}	& 3.52\\
                        
\midrule
\multirow{4}{*}{Task 6} & Phi-4 Mini        & \textbf{1.227} & 1063.4 & \textbf{7.296}  & \underline{6.0} & \textbf{3648.00}  & \textbf{2.59} \\
                        & Qwen-3       & 6.904 & \underline{5416.7} & 27.509 & 1.4 & 13754.50 & 3.04 \\
                        & Menta & 5.741 & \textbf{5851.4} & \underline{19.883} & 1.3 & \underline{9941.50}  & \underline{2.94} \\
                        & StableLM	& \underline{2.51}	& 36.9	& 23.1	& \textbf{6.4}	& 11500.00	& 3.53\\
\bottomrule
\end{tabular}
\end{table}

\end{document}

%% file: math_commands.tex
\usepackage{amsmath,amsfonts,bm}

\def\eqref#1{equation~\ref{#1}}

\def\1{\bm{1}}

\DeclareMathAlphabet{\mathsfit}{\encodingdefault}{\sfdefault}{m}{sl}
\SetMathAlphabet{\mathsfit}{bold}{\encodingdefault}{\sfdefault}{bx}{n}

